\documentclass[11pt]{article}
\usepackage[preprint]{acl}
\usepackage{times}
\usepackage{latexsym}
\usepackage[T1]{fontenc}
\usepackage[utf8]{inputenc}
\usepackage{microtype}
\usepackage{inconsolata}
\usepackage{graphicx}
\usepackage{booktabs}
\usepackage{multirow}
\usepackage{amsmath,amssymb}
\usepackage{xcolor}
\usepackage{url}

\title{Presentation, Not Mechanism: A Render Confound in Deprecation-Aware Memory Evaluation}

\author{%
  \bfseries
  Zhaoyang Jiang\textsuperscript{1} \quad
  Zhizhong Fu\textsuperscript{3} \quad
  Zicheng Li\textsuperscript{1} \quad
  Yunsoo Kim\textsuperscript{4} \\[0.3em]
  \bfseries
  Jiacong Mi\textsuperscript{1} \quad
  Xuanqi Peng\textsuperscript{1} \quad
  Fei Teng\textsuperscript{2} \quad
  Honghan Wu\textsuperscript{1}\thanks{Corresponding author.} \\[0.5em]
  \normalfont
  \textsuperscript{1}School of Health \& Wellbeing, University of Glasgow, Glasgow, UK \\
  \textsuperscript{2}Department of Respiratory and Critical Care Medicine, Shanghai Sixth People's Hospital, \\
  Shanghai Jiao Tong University School of Medicine, Shanghai, China \\
  \textsuperscript{3}School of Life Science and Technology, \\
  University of Electronic Science and Technology of China, Chengdu, China \\
  \textsuperscript{4}Institute of Health Informatics, University College London, London, UK \\[0.4em]
  \{3167645J, 3131960P, 3222974L\}@student.gla.ac.uk, \quad j.mi.1@research.gla.ac.uk, \\
  Honghan.Wu@glasgow.ac.uk, \quad tengfeifaw@163.com, \quad zhizhong.fu@std.uestc.edu.cn, \\
  yunsoo.kim.23@ucl.ac.uk
}

\newcommand{\rl}{RevisionLedger}
\newcommand{\grw}{GraphRAG+abstain}

\newcommand{\abstain}{\textsc{Abstain}}
\newcommand{\sys}[1]{\textsc{#1}}

\begin{document}
\maketitle

\begin{abstract}
AI systems increasingly retrieve from records that revise themselves: issue threads, encyclopedic histories, policy logs, and long conversations. The challenge is not only finding relevant evidence, but deciding which claims remain in force, which were superseded, and when to abstain. Structured memories promise to solve this with typed edges, temporal updates, and conflict status, yet evaluations often change mechanism and prompt presentation together. We study this as Evidence-State Revision, comparing flat retrieval, coarse edge invalidation, and fine-grained \rl{} on 2{,}907 high-agreement questions from GitHub, multi-repo issue histories, Wikipedia, and DyKnow-style temporal streams. A render-matched control (same layout, deprecation disabled) reveals the central confound: when a value is changed and later restored, \rl{} appears to beat a flat baseline by $+0.182$, but almost all the gain comes from easier presentation; the fine-grained mechanism residual is indistinguishable from zero ($+0.021$ to $+0.025$ across two judge families). After presentation is controlled, coarse invalidation is the only mechanism that pays for current-state queries, beating the fine ledger by $0.084$; the same query-sufficiency principle says provenance mainly needs retained invalidated evidence, not richer typing. Memory evaluations should hold render fixed, and deprecation-aware systems should deploy the coarsest retained state that covers their queries.
\end{abstract}

\section{Introduction}

Retrieval systems increasingly answer questions over records that revise themselves: software issue threads, encyclopedic histories, policy logs, incident reports, and long-running conversations. A maintainer may first recommend a workaround, later replace it with a patch, and later still qualify that patch by platform. In such streams, we call a claim \emph{in force} when it is the value a system should use after later corrections, refinements, and supersessions. The right answer is therefore not always the newest sentence or the majority of retrieved passages; it is the value that remains live in the revision history. This matters because stale state is a high-impact failure mode for coding assistants, support agents, and analytic systems that must know when to answer, when to cite a prior value, and when to abstain. The central question is not simply whether memory should be ``structured,'' but what structure is actually necessary for the query being asked.

Existing systems answer pieces of this question but not the selection problem. Flat RAG and GraphRAG-style systems retrieve and group evidence, yet mostly treat retrieved facts as additive: old and new claims are placed together as usable context, with no explicit decision about which one is still in force \citep{edge2024graphrag,gutierrez2024hipporag}. Agent memories store evolving facts, but often leave supersession implicit in summaries or embeddings \citep{packer2023memgpt,xu2025amem,chhikara2025mem0}. Temporal graph memories such as Zep/Graphiti invalidate old edges when a new one supersedes them, giving a coarse notion of deprecation \citep{rasmussen2025zep}. Knowledge-conflict benchmarks study contradictions among passages or between retrieved and parametric knowledge \citep{xu2024knowledge,hou2024wikicontradict,ming2024faitheval,li2024contradoc,xie2024adaptive}, but usually diagnose conflict at answer time rather than as an ingestion-time state update. The design space is therefore unsettled: should a revising stream use ordinary retrieval, a conflict-aware merger, a supersession-aware store, or a full typed ledger?

This design question has an evaluation trap. Structured memories often change both the mechanism they implement, such as typed edges or invalidation, and the presentation they show the answer model, such as cleaner ordering and explicit fields. If a ledger beats a flat baseline, the answer LLM may be benefiting from a better-formatted prompt rather than a better memory mechanism. A render-matched evaluation is therefore part of the method, not a cosmetic control: without it, we may recommend an elaborate architecture when a simpler state representation, rendered well, would have worked.

We study this problem as \emph{Evidence-State Revision} (ESR): given an evolving evidence stream, maintain enough state to answer what currently holds, what was superseded, and when no single answer is justified. We organize streams by conflict locality: within a passage, across passages, or across time for the same entity attribute. The temporal case is distinctive because it requires a deprecation marker saying that a later event has made an earlier atom no longer live. On ESR-Bench, we compare a GraphRAG-style baseline with no deprecation marker (\grw), a coarse edge-invalidation store in the Graphiti/Zep style, and \rl{}, a fine-grained ledger with relation types and unresolved status. Our hypothesis is query-sufficiency: current-value (snapshot) questions need only live/dead deprecation, provenance needs retained invalidated evidence, and fine relation types are useful only when the query asks for relation or status itself.

\begin{figure*}[t]
  \centering
  \includegraphics[width=\linewidth]{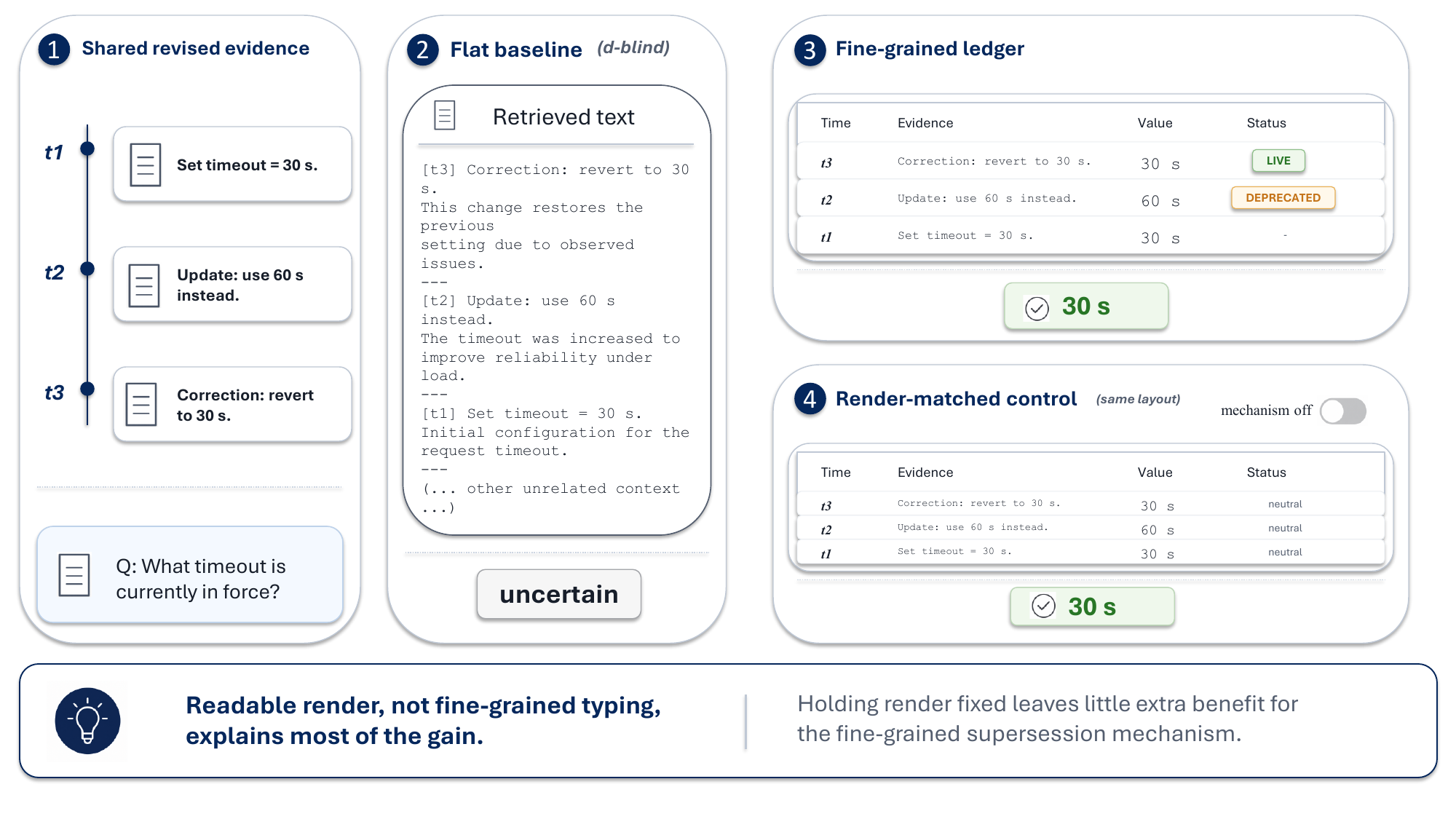}
  \caption{\textbf{The render confound.} In a reverted-revert stream ($30\text{s} \to 60\text{s} \to 30\text{s}$), \rl{} answers the current value while the flat baseline answers \emph{uncertain} (an abstention). A render-matched control keeps the ledger layout but disables deprecation labels; it still answers $30\text{s}$. Thus the apparent ledger gain can come from readable, time-ordered presentation rather than fine-grained supersession typing; \S\ref{sec:revrev} quantifies the near-zero residual mechanism effect.}
  \label{fig:confound}
\end{figure*}

Our experiments validate this view on 2{,}907 high-agreement ESR-Bench questions over GitHub, multi-repo issue histories, Wikipedia revisions, and DyKnow-style temporal streams. On reverted-revert questions, where a middle claim is superseded and the current value returns, \rl{} appears to beat a flat baseline by $+0.182$. A render-matched control shows that almost all of this gain is presentation; the fine-grained mechanism residual is indistinguishable from zero ($+0.021$ to $+0.025$ across two judge families). Coarse invalidation is the mechanism that pays for current-state queries, and on the clean setting it matches or beats the fine ledger ($\Delta_{\rl-\text{Graphiti}}=-0.084$). Provenance shows the dual lesson: a coarse store that discards invalidated edges fails because the prior value is absent, while a retaining coarse store recovers most of the apparent fine-ledger advantage. The practical recommendation is thus compact: hold render fixed when evaluating memory architectures, and deploy the coarsest retained state that covers the query mixture.

Code is available at \url{https://github.com/Anonymous-Awesome-Submissions/ESR-pipeline}. Privacy-minimized ESR-Bench artifacts are available at \url{https://github.com/Anonymous-Awesome-Submissions/ESR-bench}; consistent with the Ethics Statement, released artifacts are derived benchmark artifacts rather than raw GitHub threads, Wikipedia revisions, or other verbatim source text.

\section{Evidence-State Revision}
\label{sec:framework}

An evidence stream is an ordered sequence $C=(e_1,\ldots,e_T)$, where each event has text, a timestamp, and optional metadata. An atom extractor maps events to tuples $(x,a,v,\pi,t,\rho)$: entity, attribute, value, polarity, time, and source role. For atoms with the same entity--attribute key $(x,a)$, ESR asks which value is currently in force, which events support it, and which earlier events have been closed by later evidence. The system may output \abstain{} when no single current value is justified.

The core state is a ledger row per key $(x,a)$: the current value, valid-since time, supporting atoms, deprecated atoms, and conflict status. Relation labels are used to update this state: \textsc{supersedes} replaces the value and deprecates prior support, \textsc{refines} adds finer support, and unresolved \textsc{contradicts} routes to \abstain{}. We assume each key holds a single current value; faceted states, where a value differs by platform or version, need a finer key and are outside the current extractor.

\begin{figure*}[t]
  \centering
  \includegraphics[width=\linewidth]{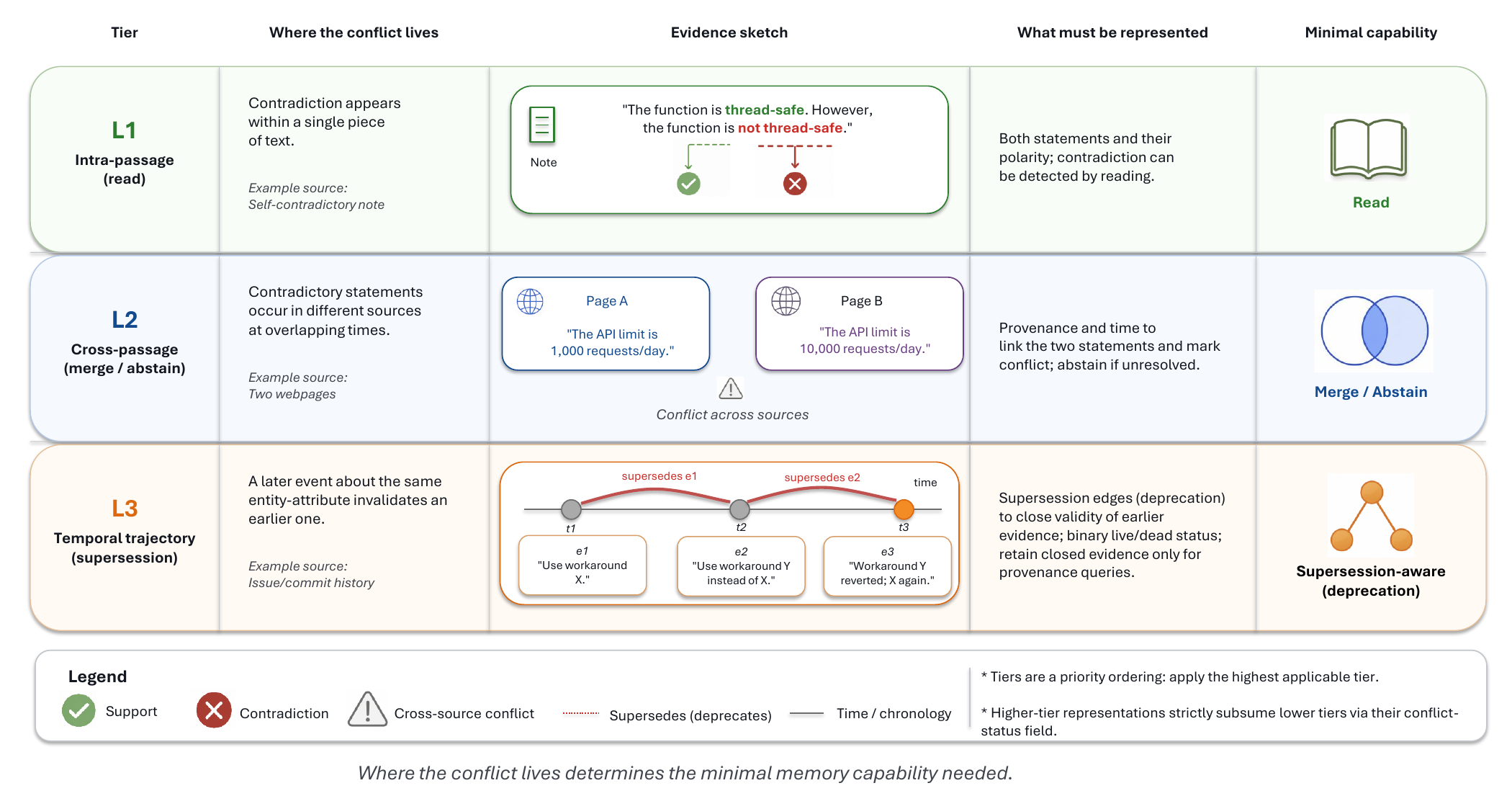}
  \caption{\textbf{Conflict locality determines memory need.} Intra-passage contradictions ($L_1$) can be settled by reading; cross-passage conflicts ($L_2$) require provenance and valid time to associate sources and abstain when unresolved; temporal entity-state revision ($L_3$) requires deprecation. Tiers are a priority ordering: the highest applicable tier's state subsumes the lower ones through its conflict-status field. The $L_3$ minimal capability is a binary live/dead deprecation marker, plus retention of invalidated evidence when prior values are queried, rather than the fine-grained relation typing our experiments find idle on current-state queries (\S\ref{sec:revrev}, \S\ref{sec:provenance}).}
  \label{fig:tiers}
\end{figure*}

The diagnostic is \emph{conflict locality}: where the conflict lives fixes the minimal primitive a memory must expose. This is a selection rule, not a claim that every stream needs the largest representation. $L_1$ needs only readable evidence, $L_2$ needs source-aware conflict status, and $L_3$ needs deprecation. Because the implemented ledger subsumes lower tiers through its conflict-status field, unresolved cross-source conflicts must still route to \abstain{} rather than be treated as temporal updates; DS \textsc{cross-source-conflict} tests this high-recall gate directly.

We write $d$ for the deprecation coordinate: whether an atom is live or deprecated. This notation separates architectures that never represent deprecation ($d$-blind) from those that do ($d$-aware), but it does not by itself justify a fine-grained ledger. Appendix~A formalizes three results we use below. Proposition~1 (P1) shows that a $d$-blind store cannot recover the current value on a symmetric subset of revising streams that look identical without $d$, giving a $50\%$ ceiling there. Proposition~2b (P2b) then shows that for snapshot, or current-state, queries the binary live/dead partition $\Pi$ induced by $d$ is a minimal sufficient statistic: once $\Pi$ is known, finer relation typing carries no additional value-relevant information (the typing benefit $b_{\text{type}}=0$). Proposition~2c (P2c) generalizes this into a query-sufficiency lattice and a retention-optimal selection rule.

The resulting recommendation is query-relative. Coarse-$d$ stores recover $\Pi$ directly through binary edge invalidation; fine-$d$ ledgers can match that partition but also pay for a typed scorer, a heavier render, and an abstain path that is brittle in current LLMs \citep{kirichenko2025abstentionbench}. Corollary~1 therefore predicts no intrinsic fine-typing gain on snapshot tasks when fine-$d$ has non-negative excess cost; the render-matched ablation of \S\ref{sec:revrev} tests this by holding presentation fixed. Fine typing has room to help only when the query reads a relation coordinate that $\Pi$ collapses, such as refinement or unresolved status, and only when that coordinate is actually recovered. Appendix Proposition~4 (P4) is used only as decision-channel bookkeeping; because P4's out-of-sample sign prediction is falsified by the controlled noise sweep in \S\ref{sec:rl_vs_graphiti}, the granularity ranking rests on P2b and Corollary~1.

\section{ESR-Bench and Memory Classes}
\label{sec:bench-systems}

ESR-Bench tests the selection criterion by holding the task fixed and varying the memory primitive. We instantiate three memory classes: \grw{} as the $d$-blind GraphRAG-class control with no deprecation marker, a coarse-$d$ edge-invalidation store in the Zep/Graphiti style, and \rl{} as the fine-$d$ ledger with relation labels and unresolved status. Comparing within and across these classes on each stratum tests when deprecation, retention, and fine relation typing are actually needed.

The benchmark is designed to expose $L_3$ temporal entity-state revision while retaining simpler $L_1$ and $L_2$ probes for falsifiability. The main ESR-Bench version contains 2{,}907 QAs drawn from GitHub issue histories, multi-repo issue histories, Wikipedia revisions, and a 109-item DyKnow temporal split. Each item stores a gold answer, supporting event ids, deprecated event ids, and a phenomenon label. An auxiliary high-noise construction (a 1{,}198-QA \texttt{huggingface/datasets} GitHub split, plus smaller multi-repo and Wikipedia splits) is used only in the appendix, not for the main claims. Public probes are outside the 2{,}907-item main version and test no-effect predictions on adjacent settings: WikiContradict \citep{hou2024wikicontradict}, FaithEval \citep{ming2024faitheval}, DyKnow-style latest-known questions, StreamingQA-style temporal QA \citep{liska2022streamingqa}, and LongMemEval-S \citep{wu2024longmemeval} where relevant.

\paragraph{Primary ESR-Bench version.}
All main experiments use DeepSeek-V4-flash with non-thinking decoding, a shared QA prompt template, two-pass agreement filtering, and a bilingual phenomenon classifier. This version contains 1{,}698 GitHub, 252 multi-repo, 848 Wiki, and 109 DyKnow QAs. The rare strata comprise 99 reverted-revert, 88 cross-source conflict, and 89 refines questions; the remaining 2{,}631 are monotonic.

\paragraph{Human validation.}
\label{sec:human-validation}
A domain expert hand-validated a stratified $N=150$ subset under the same bilingual rubric the LLM labeller saw (per-stratum agreement in Appendix). This check supports the two gold assumptions used later. First, the human-confirmed canonical subset preserves the reverted-revert gap ($\Delta_{\rl-\grw}=+0.304$ on the $25$ canonical reverted-revert items, $n_{\text{paired}}{=}112$), and \S\ref{sec:revrev} shows that this larger canonical gap is still driven mainly by render rather than mechanism. Second, on cross-source-conflict items, human validation confirms that abstention is the correct gold behavior in $94.7\%$ of cases, supporting the abstention-channel analysis in \S\ref{sec:ds-channels}. Overall phenomenon-label agreement is Cohen's $\kappa=0.459$ ($95\%$ CI $[0.366,0.553]$); disagreement is asymmetric because the LLM over-labels \textsc{reverted-revert}, half of which are canonically \textsc{monotonic}.

\begin{table}[t]
\centering
\scriptsize
\setlength{\tabcolsep}{3pt}
\begin{tabular}{@{}p{0.86in}rrp{0.94in}@{}}
\toprule
Evaluation set & Items & Families & Role \\
\midrule
GitHub & 1{,}698 & 5 & Primary \\
Wiki & 848 & 4+partial & Primary \\
MultiRepo & 252 & 5 & Primary \\
DyKnow & 109 & 4 & Primary \\
\midrule
Primary total & 2{,}907 & n/a & Primary \\
\midrule
Auxiliary ESR split & 1{,}198 & 5 & Auxiliary \\
\bottomrule
\end{tabular}
\caption{Main and auxiliary ESR-Bench evaluation sets. The four primary sets sum to 2{,}907 QAs. ``Families'' counts paired answer-LLM families; Wiki includes partial Mistral-7B rare-stratum follow-ups, while DyKnow omits Mistral-7B. Paired bootstrap uses $B=10{,}000$ unless otherwise noted.}
\label{tab:eval}
\end{table}

\rl{} instantiates the fine-$d$ $L_3$ architecture: a frozen LLM extracts atoms from a closed attribute inventory, a rule scorer labels co-keyed pairs, the ledger operator maintains current/deprecated sets and unresolved status, and the answer generator sees only the rendered ledger and an event-id index. This isolates whether a richer state representation helps the same frozen LLM answer more selectively.

The primary $d$-blind control is \grw{}: same extracted atoms, timestamps, and abstain instruction, but no supersession edges or bitemporal state update. Additional baselines cover three failure modes: retrieval-only systems (\sys{rag-top-k}, direct long context), time-only heuristics (latest-only, timestamp reranking, temporal chain-of-thought), and structure without deprecation (\sys{graphrag-lite}, metadata-only). Headline semantic-correctness numbers are judged by DeepSeek-V3 \citep{deepseekai2024deepseekv3}; abstain calibration is independent of this semantic-correctness judge because it compares binary abstain decisions to gold abstain labels. We report paired bootstrap intervals (percentile, $B{=}10{,}000$) preserving (question id, answer family) pairs.

\section{Main Results}
\label{sec:main-results}

\subsection{Most Reverted-Revert Gain Is Render, Not Fine Typing}
\label{sec:revrev}

The central prediction concerns reverted-revert items: a middle event appears to contradict the current answer but is itself superseded later. \grw{} can detect conflict, but without the supersession edge it often abstains; the fine-$d$ ledger deprecates the middle event and answers with the surviving value. The natural reading of any ledger advantage here is that the deprecation mechanism is doing the work. We show that reading is largely wrong.

\paragraph{The headline gap.} Under the DeepSeek-V3 judge, \rl{} beats \grw{} on the primary reverted-revert slice by $+0.182$ (95\% CI $[+0.134,+0.230]$, $n_{\text{paired}}{=}439$), with positive deltas on GitHub, Wiki, and DyKnow and a tie on the tiny MultiRepo slice. The coarse-$d$ edge-invalidation store beats \grw{} by even more, $+0.266$. So on clean supersession the load-bearing primitive is plainly $d$-awareness, not fine typing; the coarse architecture is the better deployment choice, and fine-$d$ \rl{} loses to it (Corollary~1). But $+0.182$ does not establish that the ledger's \emph{deprecation mechanism}, rather than its presentation, earns the advantage over the $d$-blind baseline.

\paragraph{Render-matched decomposition.} To separate mechanism from presentation we add a render-matched control: the fine-$d$ ledger's exact structured layout, but with the deprecation mechanism switched off, forcing every co-keyed relation to \textsc{same-state} so there are no supersedes, no contradicts, and no unresolved status (a $d$-blind system with the ledger's layout). All four systems are judged uniformly by the gold DeepSeek-V3 judge \citep{deepseekai2024deepseekv3} and by an out-of-family MiniMax-M2.5 judge, and the decomposition is stable across both (Table~\ref{tab:render_decomp}). Under DeepSeek-V3 the reverted-revert gain ($+0.184$, matching the $+0.182$ headline) splits into a \emph{render} term of $+0.159$ (render-matched control $-$ \grw{}) and a $d$-\emph{mechanism} residual of only $+0.025$ (\rl{} $-$ render-matched control), the latter indistinguishable from zero; MiniMax agrees. On the $25$ human-confirmed canonical reverted-revert items, the split is starker: the render component carries essentially the entire $+0.304$ canonical total while the mechanism residual remains indistinguishable from zero ($+0.036$ $[-0.027,+0.098]$). The fine-$d$ deprecation machinery adds no measurable accuracy even on the cleanest items; almost all of the ledger's apparent edge is that its structured layout is easier for the answer-LLM to read than \grw{}'s flat entity-attribute groupings. The one mechanism that does pay is coarse-$d$ invalidation ($+0.087$ over the render-matched control), a real but coarse deprecation effect. This is the central correction of this paper: a structured-memory comparison that does not hold render fixed will misattribute a presentation effect to its mechanism.

\begin{table}[t]
\centering
\footnotesize
\setlength{\tabcolsep}{3.5pt}
\begin{tabular}{@{}lcc@{}}
\toprule
Term & DeepSeek-V3 [95\% CI] & MiniMax \\
\midrule
Render & $+0.159$ $[+0.114,+0.207]$ & $+0.164$ \\
Fine-$d$ mech. & $+0.025$ $[-0.005,+0.057]$ & $+0.021$ \\
Coarse-$d$ mech. & $+0.087$ $[+0.046,+0.130]$ & $+0.098$ \\
\bottomrule
\end{tabular}
\caption{\textbf{Most of the fine-ledger gain is render.} Render-matched decomposition (Render $=$ render-matched control $-$ \grw{}; fine-$d$ mech.\ $=$ \rl{} $-$ control; coarse-$d$ mech.\ $=$ Graphiti $-$ control). The render term dominates; the fine-$d$ mechanism residual is indistinguishable from zero (CI crosses $0$); only coarse-$d$ invalidation adds a real but small effect. The out-of-family MiniMax-M2.5 judge agrees, so the split is not a single-judge artifact.}
\label{tab:render_decomp}
\end{table}

\paragraph{Why Fine Typing Is Idle on Snapshot and Where It Pays.} Proposition~2b makes the render result coherent and predicts it generalizes: the typed mechanism residual (\rl{} $-$ render-only) is statistically zero on \emph{all four} strata (Table~\ref{tab:query-dependence}). On snapshot queries this is forced: $\Pi$ is sufficient, so $b_{\text{type}}=I(v^\star;R\mid\Pi,\text{values})=0$. But the theorem is \emph{query-relative}: where the answer reads a relation coordinate that $\Pi$ collapses, \emph{recovering} it pays, as shown by a learned scorer on \textsc{refines} and a strict-prompt decode on \textsc{cross-source-conflict} (Table~\ref{tab:query-dependence}). There is no analogue on snapshot queries, where the signal is provably absent from $R$ beyond $\Pi$. Typed structure pays only when the query reads a coordinate of $R$ that $\Pi$ collapses \emph{and} that coordinate is recovered: the bitemporal current-value/provenance distinction, made quantitative.

\begin{table}[t]
\centering
\scriptsize
\setlength{\tabcolsep}{4pt}
\resizebox{\columnwidth}{!}{%
\begin{tabular}{@{}lccc@{}}
\toprule
Stratum (query) & Coord.\ in $R$? & Typed mech.\ residual & After recovery \\
\midrule
reverted-revert (snapshot) & no ($\Pi$ only) & $+0.025$ (n.s.) & n/a \\
monotonic (snapshot) & no ($\Pi$ only) & $-0.003$ (n.s.) & n/a \\
\textsc{refines} (granularity) & yes & $+0.009$ (n.s.) & $+0.147$\textsuperscript{a} \\
\textsc{csc} (status) & yes & $-0.008$ (n.s.) & $+0.152$\textsuperscript{b} \\
\bottomrule
\end{tabular}}
\caption{\textbf{Fine typing pays only when the queried coordinate is recovered.} The typed-mechanism residual is \rl{} minus the render-only control under the same scorer and prompt; it is statistically zero on every stratum. Where the answer reads a coordinate of $R$ that $\Pi$ collapses, \emph{recovering} that coordinate then pays: \textsuperscript{a} a learned scorer on \textsc{refines}, \textsuperscript{b} a strict-prompt decode on \textsc{csc}. There is no analogue on snapshot queries, where $b_{\text{type}}=0$ makes the signal provably absent.}
\label{tab:query-dependence}
\end{table}

\subsection{\texorpdfstring{Controlled Noise Does Not Recover a Fine-$d$ Advantage}{Controlled Noise Does Not Recover a Fine-d Advantage}}
\label{sec:rl_vs_graphiti}
When, if ever, does fine-$d$ overtake coarse-$d$? On the clean primary reverted-revert slice the coarse store beats \rl{} ($\Delta_{\rl-\text{Graphiti}}=-0.084$ $[-0.128,-0.041]$). An earlier reading inferred a sign flip from the contrast with a noisier auxiliary split (fine-$d$ led by $0.090$), but that contrast confounds extractor noise with annotator identity and relation-type mix. We replace it with a controlled relation-noise sweep: holding dataset, judge, and answer model fixed, we corrupt each predicted relation to a random other label with probability $p$, identically in both arms. Across $p\in\{0,\dots,0.5\}$ and five families ($n{=}445$/level), the arms tie at clean signal on the sweep coverage ($\Delta=-0.016$, n.s.) and every positive noise level still has a coarse-$d$ point-estimate lead (Appendix Figure~\ref{fig:noise}; Appendix~\ref{sec:noise}). The relation-noise curve is neither monotone in $p$ nor sign-flipping, falsifying Appendix Proposition~4's monotone-crossover prediction (C3) on the axis it names.

We also test the \emph{extraction} axis directly, dropping each atom with probability $p$ (recall loss, the dominant extractor failure) identically in both arms. With the same $n{=}445$/level, coarse-$d$ still leads or ties at every tested level (Appendix Table~\ref{tab:noise-curve}). Neither the relation-scoring nor the extraction axis yields a tested regime where fine grain wins; we do not claim every conceivable noise process is covered. Coarse $d$-awareness is the right granularity here, the simplicity-under-clean-signal pattern \citep{laitenberger2025strongerrag} specialized to bitemporal memory.

\subsection{Provenance Requires Retention, Not Fine Typing}
\label{sec:provenance}
\begin{figure}[t]
  \centering
  \includegraphics[width=\columnwidth]{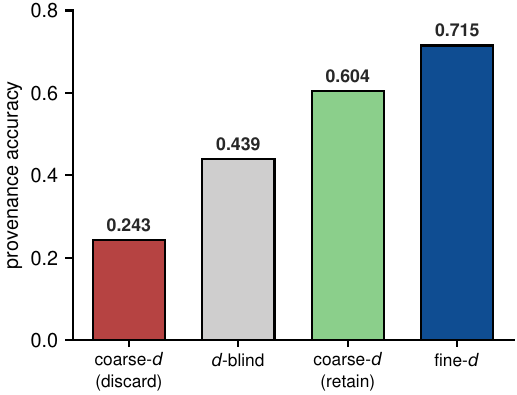}
  \caption{\textbf{Provenance needs retention, not typing.} Accuracy on provenance queries (``what value held just before?'') from retained state only, pooled over five answer-LLM families (DeepSeek judge, $n{=}4{,}895$). A coarse store that \emph{discards} closed edges cannot answer the prior-value query; \emph{retaining} them, whether coarse or fine, restores answerability. The fine-$d$ bar is therefore an apparent edge over discard, not evidence that fine relation typing is the causal ingredient.}
  \label{fig:provenance}
\end{figure}

The provenance result is a retention threshold, not a fine-typing win. The sufficiency lattice (Proposition~2c) is two-sided: if the binary partition is sufficient for snapshot queries, it must be \emph{insufficient} for queries that read what it discards. We test the sharpest such case, a \emph{provenance} query: ``what value was in force immediately before the current one?'' We construct 979 such queries whose gold prior value comes from the benchmark's own \texttt{gold\_deprecated\_event\_ids} (independent of any system's scorer), and present each architecture with only its retained \emph{state} (no raw-event fallback). Pooled over five answer-LLM families ($n{=}4{,}895$, DeepSeek judge; an out-of-family MiniMax-M2.5 judge reproduces the ordering), a coarse-$d$ that \emph{discards} closed edges fails, while any store that \emph{retains} closed edges with their \texttt{invalid\_at} windows, whether coarse or fine, clears the threshold (Figure~\ref{fig:provenance}). Fine-$d$ is highest in this state-only diagnostic, but the visual ordering should be read as answerability once prior state is retained, not as evidence that relation labels are the cause.

The reason is information-theoretic: a store that discards superseded values cannot answer because the prior value is \emph{absent} from its state. Retention repairs that missing state. This is the dual of \S\ref{sec:revrev}: there fine-$d$'s extra state was wasted ($b_{\text{type}}=0$); here the missing state is fatal, but what repairs it is retention, not granularity.

\paragraph{What this does and does not credit to fine-$d$.} The apparent fine-$d$ provenance win is $+0.472$ $[+0.455,+0.489]$ over a discarding coarse store, whose accuracy is $0.243$. But real edge-invalidation memories (Zep/Graphiti) do not discard: they stamp a superseded edge \texttt{invalid\_at} and keep it. A faithful retention-coarse-$d$ keeps closed edges with their windows but uses \emph{no} typed relations; it scores $0.604$, closing $+0.361$ $[+0.344,+0.377]$ of that gap. Most of the apparent provenance win was therefore a discard \emph{modeling choice}, not a property of the coarse-$d$ class; crediting it to typing would be the snapshot-side error in reverse. The honest residual is $\mathrm{fine}-\mathrm{retain}=+0.111$ $[+0.097,+0.125]$, significant but a fraction of $+0.472$, and plausibly itself \emph{presentation}, since fine-$d$ hands the reader a pre-ordered prior history while retention-coarse-$d$ hands back unordered timestamped edges to sort. The out-of-family MiniMax-M2.5 judge agrees that the residual is small, $+0.063$ on its shared subset.

\paragraph{Deployment renders can reverse the provenance residual.} The provenance effect is memory-constrained: using each system's \emph{deployed} render (raw-event text appended) reverses the sign, with fine-$d$ at $0.311$, coarse-$d$ at $0.546$, $d$-blind at $0.610$, and $\mathrm{fine}-\mathrm{coarse}=-0.236$ $[-0.250,-0.221]$, because the heavier ledger render now hurts the reader, the same render effect as \S\ref{sec:revrev} with opposite sign.

We therefore do not claim typed relations decisively win provenance; we claim retention is necessary, coarse retention captures most of it, and even the residual narrows to presentation. The regime where retention matters at all is the one memory architectures are built for: compressed state under a retention budget, such as on-device, privacy-constrained, or cost-capped deployments that cannot keep the raw stream. The retention threshold replicates on external TempLAMA, with Wikidata gold, exact-match, and no LLM judge, where a discarding store falls to floor accuracy and retention recovers $+0.514$; Appendix~\ref{sec:templama} gives the full ordering, and DyKnow $n{=}11$ concurs.

\subsection{Boundary Strata: Typed Gains Require Coordinate Recovery}
\label{sec:ds-channels}
The \textsc{csc} and \textsc{refines} recovery gains ($+0.152$, $+0.147$; Table~\ref{tab:query-dependence}) are not render-matched and so upper-bound the mechanism contribution. The same strict-abstain prompt does \emph{not} move \grw{} ($+0.006$, n.s.), and a learned scorer flips \textsc{refines} from $-0.086$ to $+0.147$, with both queries reading a coordinate that $\Pi$ collapses (Proposition~2b; Appendix Table~\ref{tab:ds-channel}).

On \textsc{monotonic}, the $\rl{}-\grw{}$ gap is larger on items carrying deprecated evidence ($+0.073$ vs $+0.049$ on the pure-no-revision half), yet even that pure half is mostly render: a \textsc{same-state} ablation already beats \grw{} by $+0.027$ with full \rl{} adding only $+0.022$, the same render-over-mechanism split as reverted-revert.

\section{Related Work}

Temporal graph memories already implement part of the state we study. Zep/Graphiti \citep{rasmussen2025zep}, for example, closes an edge's validity window when a newer edge supersedes it, which is a coarse-$d$ implementation in our taxonomy. Our concern is not whether such systems are useful, but how their gains are attributed. Published comparisons of structured memory against flat baselines often give the structured system a distilled, reorganized fact set while giving the baseline raw chunks or the full transcript (Zep vs.\ MemGPT/full-context \citep{rasmussen2025zep}; Mem0 vs.\ full-context and RAG \citep{chhikara2025mem0}). Because mechanism and prompt presentation then co-vary, and because LLMs are sensitive to context layout and length \citep{liu2024lost,cuconasu2024power}, those comparisons do not by themselves establish which part of the architecture did the work. Our render-matched control isolates that attribution question.

Recent memory systems map onto the design points our criterion distinguishes, rather than forming a single leaderboard. Write-time reconciliation and supersession handlers such as WorldDB and Zep/Graphiti sit at the coarse-$d$ point \citep{worlddb2026,rasmussen2025zep}; our snapshot results predict that this point is already sufficient, provided invalidated evidence is retained when provenance is queried. Typed temporal memories add relation structure \citep{memanto2026}, while versioned-graph accounts give belief-revision semantics for fixed memory graphs \citep{park2026graphnative}; our criterion assigns that extra structure value only for queries that read type or status coordinates. Routing systems choose among memory paths inside a fixed architecture \citep{memrouter2026,selroute2026}, while privacy-aware forgetting work varies the retention budget on the storage frontier \citep{alqithami2025forgetful}. ArbGraph \citep{arbgraph2026} targets pre-generation conflict arbitration, closer to our $L_2$ unresolved-conflict tier than to $L_3$ temporal supersession. Their mechanisms are complementary; the missing layer is selection and attribution: when a retained statistic is necessary, when it is idle, and whether an observed gain is mechanism or render.

Knowledge-conflict benchmarks study contradictions between context and parametric memory \citep{xu2024knowledge,xie2024adaptive}, between retrieved passages \citep{hou2024wikicontradict,li2024contradoc}, within passages \citep{ming2024faitheval}, and by cause \citep{conflictbank2024}. Our tiers are orthogonal to the residence-based taxonomy of \citet{xu2024knowledge} (context vs.\ parametric vs.\ intra-memory): we partition retrieved evidence by spatiotemporal locality. The $L_3$ case, where resolution lives in a supersession edge and an earlier claim remains useful for provenance, has no clean analogue in those benchmarks. ESR-Bench is therefore a vehicle for testing architecture selection under revision, not primarily another conflict benchmark.

Our path-recall decomposition (Appendix Propositions~4--5) is closest in spirit to empirical RAG-error taxonomies \citep{leung2025rag_taxonomy}, but uses oracle substitutions to localize which pipeline stage loses the decision-relevant signal. This is a diagnostic use of the causal-probing style \citep{pearl2009causality,vig2020investigating}, not a controlled-direct-effect claim. The state primitives are older than agent memory: belief update \citep{katsuno1991difference,alchourron1985logic,konieczny2002merging} and valid-time databases \citep{snodgrass1999developing,date2002temporal} already distinguish current state from history, while selective classification and conformal risk control frame the unresolved-conflict abstention channel \citep{pugnana2023auc,angelopoulos2024conformal,xu2025scrc}.

\section{Conclusion}

Evidence-State Revision turns memory selection into a query-sufficiency question. Whether a task needs deprecation-aware memory is governed by where the conflict lives and what the query asks: P1 gives streams on which no $d$-blind architecture beats a $50\%$ ceiling, P2b shows that binary live/dead state is sufficient for current-state queries, and P2c generalizes these facts into a query-sufficiency lattice with a retention-optimal selection rule. The experiments match this lattice. On snapshot questions, coarse-$d$ matches or beats fine-$d$, and the render-matched ablation attributes the fine ledger's apparent edge to layout; on provenance, the missing primitive is \emph{retention}, not finer relation typing. The retention conclusion also replicates on external Wikidata-gold TempLAMA with exact-match scoring and no LLM judge (Appendix~\ref{sec:templama}). The practical rule is exactly the lattice's selection rule: deploy the coarsest retained state that covers the query mixture.

Two consequences follow. First, memory-architecture evaluations should hold \emph{render} fixed, because a structured store's apparent edge over a flat baseline can be a reading-ease effect rather than mechanism, as our own headline number turned out to be. We specify the render-matched control as a reusable protocol (\S\ref{sec:repro}) so future comparisons can separate layout from mechanism. Second, the ``best memory system'' question is ill-posed without a stream prior and a query distribution. Our criterion catches two opposite errors: \emph{over-building} typed history for current-state queries, where P2b makes it idle, and \emph{under-building} by retrieving a revising stream with no supersession tracking, capped by P1's ceiling. Concurrent agent-memory work \citep{park2026graphnative,memrouter2026} optimizes \emph{inside} a fixed class; ESR asks which class is necessary in the first place.

\section*{Ethics Statement}

\textbf{Data minimization.} The benchmark is built from public GitHub issue threads and Wikipedia revisions, which can carry usernames and incidental personal information even when publicly accessible. We therefore release only \emph{derived, privacy-minimized} material: event IDs, atom-level extractions, gold labels, and source URLs, but never raw thread or revision text. Before release we strip verbatim source spans and actor fields, scrub residual user handles, bare contributor names, and personal URL paths from free-text and extracted fields, and scan for personal data and credentials (emails, API keys and tokens, IP addresses, credential-like identifiers). Because the source URLs let auditors re-fetch the upstream text, the release is re-traceable rather than fully anonymized; we remove items from the index on upstream deletion, license change, or removal request. The released examples are derived extractions of public content, not verbatim user text.

\textbf{Dual use and retention.} Two risks are intrinsic to the mechanism, not the data. First, supersession-aware state tracking is dual-use: applied to people rather than software facts, the same machinery profiles how an individual's stated attributes change over time, so deployments over personal data warrant access controls. Second, the retention our provenance results favour for accuracy is in tension with data minimization and the right to be forgotten: a store that keeps invalidated evidence remembers more about prior states; deployments should bound retention and propagate deletions into the ledger.

\textbf{Use and validation.} The method can improve audit trails by surfacing deprecated evidence, but rule-derived supersession labels should not be read as human ground truth; we report unresolved-conflict abstention and deprecated-evidence provenance with every ledger row. Human validation (\S\ref{sec:human-validation}) was performed by a domain expert on public technical and revision records, not on private individuals, and releases no annotator personal data.

\section*{Limitations}
\label{sec:limits}

\textbf{Generalization.} Our central reading has two separable parts: presentation explains much of the fine-ledger's apparent advantage, and retention, not relation granularity, is the state a discarding store lacks for provenance; coarse deprecation remains the load-bearing mechanism for non-monotone current-state queries. This claim is established on a single architecture family (a fine-$d$ ledger, a coarse-$d$ invalidation store, and a $d$-blind baseline sharing one atom pipeline) and primarily one self-built benchmark; TempLAMA gives an external check for the retention result, but not for the high-clutter prose setting where the render confound is most consequential. The natural cross-family check---an independently implemented graph memory tested against a ledger under a render-matched control---remains future work.

\textbf{LLM judges and label noise.} Semantic-correctness scoring relies on LLM judges, and the phenomenon labels carry only moderate human agreement (Cohen's $\kappa=0.459$), so per-stratum effects are noisy. We mitigate this with two independent strong-judge families that agree on the near-zero fine-$d$ mechanism residual (\S\ref{sec:revrev}) and by replicating the retention threshold on external TempLAMA (Wikidata gold, exact-match, no LLM judge; Appendix~\ref{sec:templama}); residual cross-judge and label noise on the rarer strata nonetheless remains.

\textbf{Oracle assumptions and faceted state.} The sufficiency results (P2a/P2b) assume exact extraction, exact co-keying, and single-valued keys; in practice our extractor misses roughly one gold-relevant event in seven (Appendix~\ref{sec:repro}, ``Extractor coverage''), so the reported gains are net of that loss. The single-valued-key assumption is the most consequential: \emph{faceted} state, where a value depends on platform or version, needs a finer composite key and falls outside the current scope, with per-facet ledgers the natural extension.

\textbf{Theory scope and an evaluation gap.} The channel benefit-cost identity's one predictive sign claim (a noise-driven fine-vs-coarse sign flip) is \emph{falsified} on our data (Appendix~\ref{sec:noise}); we therefore use it only as diagnostic bookkeeping, and the granularity conclusion rests on the render-matched decomposition (\S\ref{sec:revrev}) and Corollary~1. This benchmark limitation is partly structural: the field lacks a public benchmark for high-clutter \emph{real} revision streams with human gold, the regime where the render confound is both measurable and most consequential. We flag this as a field-level evaluation gap that this work motivates.

\bibliography{references}

\appendix
\section{Proofs}

\begin{figure*}[t]
  \centering
  \includegraphics[width=\linewidth]{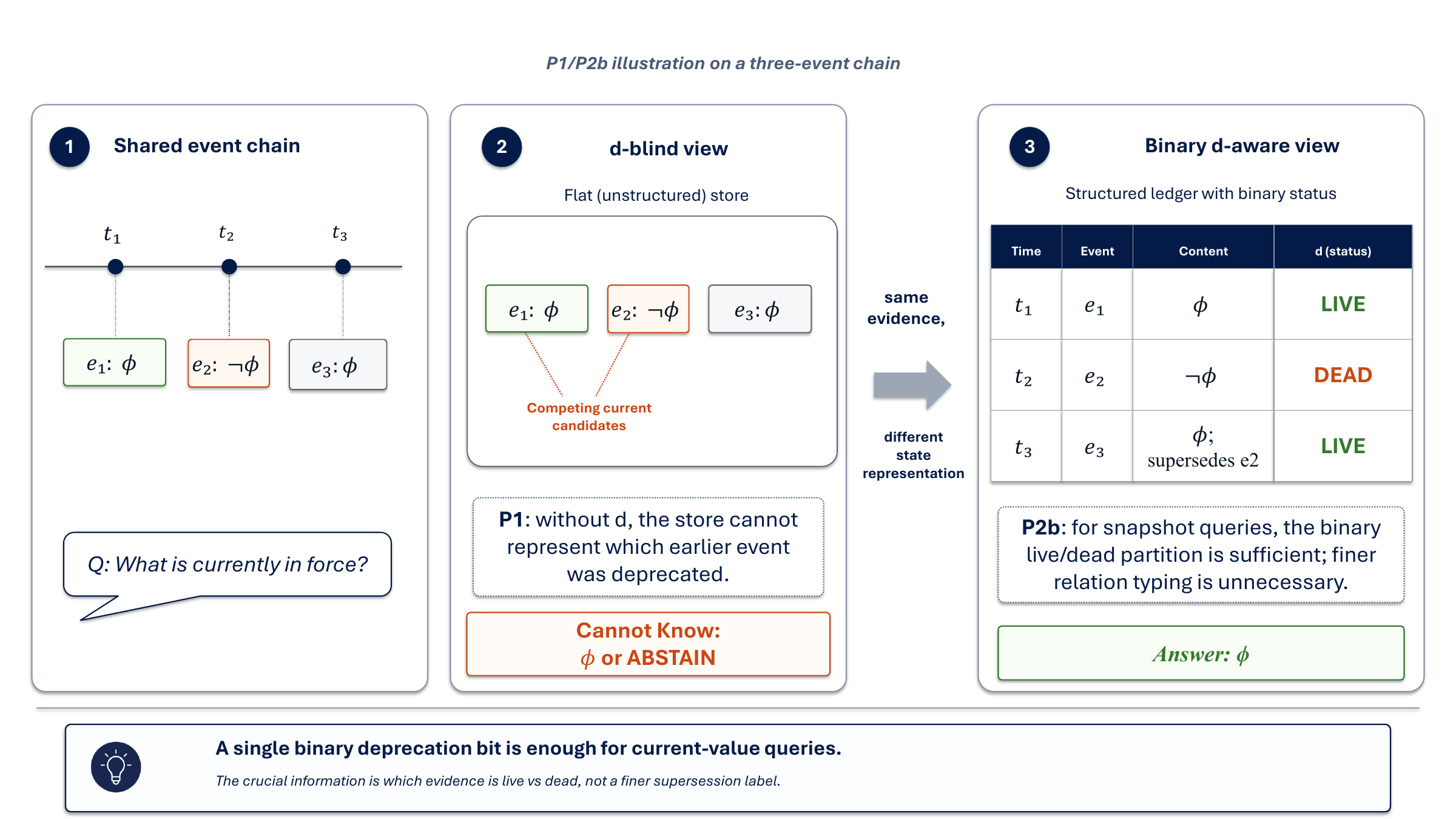}
  \caption{\textbf{A binary live/dead marker is enough for current-value queries} (P1/P2b). On the three-event chain $e_1{:}\phi$, $e_2{:}\neg\phi$, $e_3$ superseding $e_2$, a $d$-blind store cannot represent which earlier event was deprecated, so it cannot distinguish this resolved revert from an unresolved conflict and must give the same answer to both surfaces (P1, necessity). A store that records only binary live/dead status marks $e_2$ dead and returns the surviving value $\phi$; finer relation typing is unnecessary for snapshot queries once the live/dead partition is known (P2b). The crucial information is which evidence is live versus dead, not a finer supersession label.}
  \label{fig:mechanism}
\end{figure*}

\subsection*{Proposition 1 (non-monotonic supersession separation).}
\textit{Setup.} Fix a key $(x,a)$ with event set $E$, each event carrying a value and a valid time $T_v$. Let $\mathcal{A}_{\textsc{ek}+T_v}$ be the class of \emph{deterministic} architectures whose decision is a function of $E$'s values and valid times \emph{only} --- not of the deprecation marker $d$. This subsumes \sys{latest-only} (output the value of $\arg\max_{e} T_v(e)$), \sys{graphrag-lite} (output some value, never \abstain{}), and \grw{} (detect value-inconsistency, then \abstain{} or pick a $T_v$-tie-breaker). The bitemporal ledger $A_{\textsc{btl}}$ additionally reads $d$ and applies $\oplus$ (\S\ref{sec:framework}). The $L_3$ ground truth for $(x,a)$ is the value supported by the maximal $d$-non-deprecated subset of $E$, or \abstain{} if that subset is value-inconsistent.

\textit{Claim (deterministic separation).} No deterministic $A \in \mathcal{A}_{\textsc{ek}+T_v}$ matches $A_{\textsc{btl}}$ on all $L_3$ instances.

\textit{Proof.} Take the surface configuration $E = \{e_1{:}\phi @ t_1,\ e_2{:}\neg\phi @ t_2,\ e_3{:}\phi @ t_3\}$ with $t_1 < t_2 < t_3$ (Figure~\ref{fig:mechanism}), and two ground-truth-distinct instances over it:
\begin{itemize}\setlength\itemsep{0pt}
\item[(I)] $d_{\mathrm{I}} = \{(e_3, e_2)\}$: $e_3$ reverts $e_2$. The maximal non-deprecated subset is $\{e_1, e_3\}$, value-consistent on $\phi$ $\Rightarrow$ $L_3$ truth $= \phi$.
\item[(II)] $d_{\mathrm{II}} = \emptyset$: no deprecation. The maximal non-deprecated subset is $\{e_1, e_2, e_3\}$, value-\emph{inconsistent} ($\phi$ vs $\neg\phi$) $\Rightarrow$ $L_3$ truth $= \abstain$.
\end{itemize}
$E$, its values, and its valid times are identical across (I) and (II); only $d$ differs. By definition, every deterministic $A \in \mathcal{A}_{\textsc{ek}+T_v}$ has $A(\text{I}) = A(\text{II})$. But the ground truths differ ($\phi \neq \abstain$), so $A$ is wrong on at least one of the two. $A_{\textsc{btl}}$ reads $d$ and applies $\oplus$, returning $\phi$ on (I) and \abstain{} on (II): correct on both. Hence no deterministic $A \in \mathcal{A}_{\textsc{ek}+T_v}$ matches $A_{\textsc{btl}}$ on the full $L_3$ class. $\square$

\textit{Why this is not vacuous.} The construction shows the gap is not closable by any $T_v$-based decision rule --- not just that a particular baseline fails. The deterministic restriction is essential and not stronger than the empirical baselines we compare against: every system in $\mathcal{A}_{\textsc{ek}+T_v}$ in our experiments (\sys{latest-only}, \sys{graphrag-lite}, \grw{}) is deterministic in its retrieval-time decision rule. A randomized extension of $\mathcal{A}_{\textsc{ek}+T_v}$ that \abstain{}s with probability $p\in(0,1)$ on detected inconsistency does not match $A_{\textsc{btl}}$ either, but for the weaker reason that it loses correctness in expectation rather than per-instance; we state and prove the deterministic case because it is the empirically relevant one and because it admits the cleaner pointwise statement. The underlying obstruction is \emph{pointwise non-recoverability}: $d$ is not a function of $(E\text{'s values},\ T_v)$, so an architecture restricted to that surface is information-theoretically blocked from recovering it pointwise. This is the same structure as an omitted-variable obstruction; the term ``identifiability'' is reserved in causal inference for parameter recovery from a distribution, which is not what is at stake here. Only access to $d$ resolves the ambiguity.

\textit{Converse (characterization).} The obstruction is confined to the non-monotone temporal part of $L_3$. Define the \emph{monotone} sub-class $L_3^{\textsf{mono}}$: for every key, the $d$-non-deprecated atom subset is value-consistent and its supersession edges respect valid-time order (no retroactive correction, no co-keyed unresolved residual). On $L_3^{\textsf{mono}}$ the snapshot value equals the latest-$T_v$ value, so \sys{latest-only} $\in \mathcal{A}_{\textsc{ek}+T_v}$ matches $A_{\textsc{btl}}$ exactly. Hence the $d$-blind class is provably sub-optimal only when a temporal deprecation fact is needed but absent from the surface view: resolved reverted-revert cases such as (I), or co-keyed temporal residual conflicts such as (II). Pure cross-source disagreement with no temporal supersession edge is an $L_2/U$-channel abstention case rather than a P1 necessity result. This is the temporal-database fact that a one-dimensional (valid-time-only) representation reconstructs a two-dimensional (bitemporal) one iff transaction time is monotone in valid time \citep{snodgrass1999developing,date2002temporal}, specialized to the QA setting. Proposition~1 therefore \emph{characterizes} which temporal items require a bitemporal store, not merely that the class as a whole does.

\textit{Render invariance.} Both $\mathcal{A}_{\textsc{ek}+T_v}$ and $A_{\textsc{btl}}$ are defined by what information their decision function reads ($\{$values, $T_v\}$ versus $\{$values, $T_v$, $d\}$), not by how that information is serialised for a downstream LLM. The separation therefore holds for \emph{any} ledger render: changing the serialisation changes the answer-LLM's per-instance success rate (the $\alpha_{\textsf{render}}$ term defined in Proposition~3 below) but not which instances the architecture can in principle distinguish. A render-format ablation thus cannot falsify this proposition or Proposition~2a; it can only move $\alpha_{\textsf{render}}$.

\textit{Empirical scope.} In the primary ESR-Bench version, instance (I) is represented by the strict \textsc{reverted-revert} stratum ($n=99$ QAs before family expansion). The empirical \textsc{cross-source-conflict} stratum ($n=88$ QAs) is an $L_2/U$ stress test: most examples ask for abstention over unresolved sources, not for temporal deprecation recovery. The architectures separate as P1 predicts on the deprecation channel (\S\ref{sec:revrev}: $d$-aware systems beat \grw{} on strict reverted-revert), while the csc outcome is governed by Proposition~4's unresolved-gate recall. We conjecture an analogous separation holds for the transaction-time-inversion sub-case (retroactive corrections where $T_v(e_a)<T_v(e_b)$ but $T_t(e_a)>T_t(e_b)$), constructible with a three-event surface in which the discriminator is $T_t$-ordering rather than $d$. We do not state this formally here because ESR-Bench-GitHub timestamps are interpolated from \texttt{created\_at}/\texttt{closed\_at}, leaving $T_t \equiv T_v$ on every item, so the case admits no empirical test in this benchmark.

\textit{Quantitative ceiling.} P1 is pointwise, but it also implies a distribution-level accuracy ceiling that no $d$-blind architecture can break.

Fix the \emph{$d$-ambiguous fibre}: surface configurations $\sigma=(\text{values},T_v)$ realisable by at least two deprecation assignments that induce distinct $L_3$ truths (the construction above shows the fibre is non-empty, and the strict \textsc{reverted-revert} stratum is built to populate it). A deterministic $d$-blind architecture emits a single answer per $\sigma$, so on the fibre its accuracy cannot exceed the Bayes ceiling of predicting the truth from $\sigma$ alone, $\mathbb{E}_\sigma[\max_y \Pr(\text{truth}=y\mid\sigma)]$; equivalently its error is at least the conditional Bayes error $\mathbb{E}_\sigma[1-\max_y\Pr(\text{truth}=y\mid\sigma)]>0$. The ceiling value depends on the prior over $d$ on the fibre; under the \emph{symmetric} prior of the (I)/(II) construction --- revert and no-revert equally likely given $\sigma$ --- $\max_y\Pr(\text{truth}=y\mid\sigma)=\tfrac12$, so no $d$-blind architecture exceeds $50\%$ on that fibre while $A_{\textsc{btl}}$ reaches $100\%$.

We can put an empirical anchor on this prior. On the human-validated subset (\S\ref{sec:human-validation}), of the $76$ adjudicated items the LLM flagged as surface-ambiguous \textsc{reverted-revert} --- the stratum built to populate the fibre --- a human found exactly $38$ to be canonical reverts and $38$ canonically non-reverts (predominantly \textsc{monotonic}), an empirical fibre prior of $0.50$. The symmetric $50\%$ ceiling is therefore not a worst-case idealization but close to the realized prior on the stratum that instantiates the fibre, so the $d$-blind Bayes ceiling on it is $\approx\tfrac12$ rather than something the data render vacuous.

The bitemporal store does not merely help: it lifts a hard $\tfrac12$ accuracy ceiling that no amount of $T_v$-reasoning, retrieval, or decoding can raise. This is the QA-recoverability analogue of information-theoretic single-pass accuracy bounds \citep{fanoqa2025}, with $d$ playing the role of the missing bits.

We emphasize the $50\%$ figure is the \emph{symmetric-fibre} ceiling, a theoretical statement; the measured pooled \grw{} accuracy on strict reverted-revert ($0.226$) sits below every $d$-aware system, which is \emph{consistent with} (but does not by itself \emph{prove}) operating under the ceiling --- $0.226<0.5$ also reflects extraction noise and the empirical $d$-prior, which we do not disentangle here. The ceiling concerns the deprecation \emph{benefit} $b$; it does not bound the implementation \emph{cost} $\kappa$ that determines whether fine-grained machinery is worth deploying (Corollary~1, and the render-matched decomposition of \S\ref{sec:revrev} showing $b$ for fine-grained typing is empirically near zero on this benchmark).

\subsection*{Proposition 2a (oracle sufficiency).}
\textit{Statement.} Suppose (i) atom extraction is exact (every event yields its true $(x,a,v,\pi,T_v)$ tuples, no spurious ones), (ii) co-keying is exact (atoms about the same entity-attribute share a key), (iii) the deprecation marker $d$ is complete and sound, (iv) each key is single-valued at the chosen granularity, and (v) deprecation edges are functional, acyclic, and applied in transaction order. Then for every key $(x,a)$ the ledger update $\oplus$ returns the current-valid-time snapshot value --- defined as the value supported by the maximal $d$-non-deprecated atom subset for that key, or \abstain{} if that subset is value-inconsistent (the current-snapshot semantics of valid-time databases, in the tradition of \citealp{snodgrass1999developing}; cf.\ \citealp{date2002temporal}).\footnote{In strict Jensen--Snodgrass usage, ``bitemporal'' requires a $(T_v, T_t)$ tuple per fact. We use ``bitemporal'' throughout the paper in the looser temporal-graph-RAG sense (valid time plus an explicit deprecation marker), retaining the term because it is established in the memory-systems literature we compare against (Zep/Graphiti, etc.).} Multi-valued facets are handled by refining the key; if the extractor fails to do so, P2a no longer applies.

\textit{Proof.} Induction on the event sequence for a fixed key. Let $M_t$ be the active set after $t$ events: all atoms not deprecated by any observed later atom under $d$; the deprecated set is its complement. The oracle $\oplus$ is defined extensionally by maintaining this active set and recomputing status from it after each transaction, not by preserving any path-dependent status flag. \emph{Base:} a single atom $a$ gives $M_1=\{a\}$, $\textsf{deprecated}=\emptyset$, and $\textsf{status}=\textsf{ok}$. \emph{Step:} assume the invariant ``support $=M_t$, deprecated $=E_t\setminus M_t$, and status is \textsf{unresolved} iff $M_t$ is value-inconsistent at the chosen granularity.'' For an incoming atom $a'$, functional acyclic deprecation and transaction-order application make the following active-set cases exhaustive:
\begin{itemize}\setlength\itemsep{0pt}
\item $a'$ deprecates one or more active atoms $B\subseteq M_t$: $\oplus$ moves every $b\in B$ to deprecated and inserts $a'$, so $M_{t+1}=(M_t\setminus B)\cup\{a'\}$. If the removed atoms carried the only conflict, recomputation clears \textsf{unresolved}; if disagreement remains, it stays unresolved.
\item $a'$ is same-state with the active atoms at the chosen granularity: $\oplus$ adds $a'$ to support, so $M_{t+1}=M_t\cup\{a'\}$ remains value-consistent.
\item $a'$ refines an active atom at finer grain: by assumption (iv), the chosen key granularity has already been refined enough for the facet to be single-valued; $\oplus$ either adds the finer-grain support to that refined key or leaves the broader key's value unchanged. The active-set invariant is preserved at the chosen granularity.
\item $a'$ is not deprecated, deprecates no active atom, and disagrees with at least one active value: $M_{t+1}=M_t\cup\{a'\}$ is value-inconsistent, so $\oplus$ sets \textsf{status}=\textsf{unresolved}.
\end{itemize}
At the last event the invariant gives exactly the snapshot value (or \abstain{}). $\square$

\textit{Grounding: the ledger is a belief-\emph{update} operator.} P2a is not an ad-hoc construction but the specialization of a known representation result. The $\oplus$ operator is not AGM \emph{revision} \citep{alchourron1985logic} --- incorporating new information about a \emph{static} world --- but Katsuno--Mendelzon \emph{update} \citep{katsuno1991difference}: each event reports that the entity's state has \emph{changed}, and $\oplus$ moves the maintained model to the post-event state. KM's representation theorem characterizes the operators satisfying update postulates U1--U8 as exactly those induced by a family of faithful partial pre-orders, each event mapping every model to the closest models consistent with it. The supersedes-edge update (install the new value, deprecate the prior support) is this closest-state update under the recency pre-order (the newest consistent atom is closest), and the unresolved-status routing is KM's disjunctive case where two incomparable closest states survive (hence \abstain{}). P2a's snapshot-correctness is then U1--U8 specialized to single-valued keys, placing the ledger in the belief-update lineage rather than re-deriving it; it also locates our contribution precisely against concurrent \citet{park2026graphnative}, who give an AGM-\emph{revision} correspondence for a \emph{fixed} versioned graph, whereas the update reading is what an evolving entity-state stream ($L_3$) actually requires and is what makes coarse-$d$ invalidation (a single faithful pre-order) already sufficient.

\subsection*{Proposition 2b (minimal sufficiency of the binary partition for snapshot queries).}
\textit{Statement.} For a key $(x,a)$, let $R(E)$ be the typed co-keyed relation labelling (the five classes of \S\ref{sec:framework}) from which the ledger derives status, and let $\Pi(E)\in\{0,1\}^{|E|}$ be the \emph{deprecation partition} it induces: $\Pi_e=1$ iff atom $e$ is live (in the maximal $d$-non-deprecated subset), else $0$. $\Pi$ is exactly the binary collapse of $R$ that maps \textsc{supersedes} (and the unresolved trigger) to a live/dead decision and merges the remaining classes. The claim is about what the \emph{finer} labelling adds \emph{beyond this collapse}: for the snapshot-value query --- return the current effective value $v^\star$ or \abstain{} --- the partition is a sufficient statistic, i.e.\ $v^\star \perp R \mid (\Pi,\text{values})$. Equivalently $b_{\text{type}}\triangleq I(v^\star; R \mid \Pi,\text{values})=0$: \emph{given the correct live/dead partition}, the five-way relation type carries no additional value-relevant information. In the deterministic query sense used here, $\Pi$ is also \emph{minimal}: any statistic sufficient for the snapshot answer must recover the live/dead distinctions needed by the read-out, and no strictly coarser collapse of that partition remains sufficient (P1 exhibits two surfaces equal in $(\text{values},T_v)$ that differ only in $\Pi$ and have different $v^\star$, so the live/dead bit cannot be dropped). The result is \emph{clarifying} rather than deep --- it is close to immediate from the snapshot semantics --- and we flag it as such; the non-trivial structure is the strict lattice of Proposition~2c.

\textit{Proof.} By P2a the snapshot value is the value supported by the live set $M=\{e:\Pi_e=1\}$, or \abstain{} if $M$ is value-inconsistent; thus $v^\star=g(\Pi,\text{values})$ for the deterministic read-out $g$. Conditioning on $(\Pi,\text{values})$ therefore fixes $v^\star$, giving $I(v^\star;R\mid\Pi,\text{values})=0$: the type label can vary freely without changing the answer once the live/dead bit and the values are known. Minimality in this deterministic query sense: any statistic that cannot recover the needed live/dead distinction merges a live and a dead atom on some surface and so changes $g$ there (resurrecting a deprecated value or killing a current one); by P1 such surfaces exist, so that statistic is not sufficient. Hence $\Pi$ --- the binary collapse of $R$ --- is the minimal sufficient statistic for $v^\star$. $\square$

\textit{Scope: this is the oracle statement.} The theorem is conditional on the \emph{correct} partition $\Pi$; it says the five-way refinement \emph{beyond} the binary live/dead collapse is value-irrelevant, not that relation scoring is unused (the collapse itself is recovered from \textsc{supersedes} edges, exactly the coarse-$d$ mechanism). It therefore predicts that an architecture which recovers $\Pi$ is oracle-optimal on snapshot queries; whether the implemented fine-$d$ scorer recovers $\Pi$ \emph{better} than coarse-$d$ is an empirical question the theorem does not settle and the render-matched ablation (\S\ref{sec:revrev}) answers in the negative. This matches the emulation premise of Corollary~1: the fine-$d$ relation alphabet collapses onto the coarse-$d$ binary alphabet, so $R_F^\star=R_C^\star$ on snapshot tasks.

\textit{What this predicts.} (i) On snapshot queries, an architecture that recovers the binary partition is oracle-optimal; finer typing adds no recoverable benefit, only the implementation cost of P3. With Corollary~1 this forces coarse-$d \succeq$ fine-$d$ on snapshot tasks --- not merely $\kappa>\beta$, but $\beta=0$. This is the theoretical content behind the render-matched decomposition of \S\ref{sec:revrev}, where the fine-$d$ deprecation mechanism contributes an insignificant residual. (ii) The sufficiency is \emph{query-relative}: it holds for value queries, not for queries whose answer is a function of $R$ itself. A \emph{refinement} query (which granularity is current?) and an \emph{unresolved-status} query (is the key in conflict?) read coordinates of $R$ that $\Pi$ collapses, so for them $\Pi$ is \emph{insufficient}. This is the bitemporal-database fact that the current snapshot answers current-value queries but not provenance/history queries \citep{snodgrass1999developing,date2002temporal}, specialised to the relation alphabet. \S\ref{sec:revrev} and Table~\ref{tab:query-dependence} test prediction (ii) directly: typed structure is worthless on snapshot reverted-revert ($b\approx 0$), while relation-coordinate queries improve only when the relevant coordinate is actually recovered, as with the learned scorer on \textsc{refines} and strict-prompt decode on \textsc{cross-source-conflict}.

\subsection*{Proposition 2c (query-sufficiency lattice: a strict hierarchy).}
\textit{Statement.} P2b is the bottom of a lattice. For a key, let the event stream induce three nested statistics: $S_{\textsf{snap}}$ = the current value (the live atom); $S_{\textsf{prov}}$ = the time-ordered value history (current plus the ordered superseded values); $S_{\textsf{type}}$ = the value history annotated with each transition's relation type (the full labelling $R$). Consider three query classes: $Q_{\textsf{snap}}$ (current value), $Q_{\textsf{prov}}$ (the value in force immediately before the current one), and $Q_{\textsf{type}}$ (was the last change a supersession or a refinement?). Then the minimal sufficient statistics are \emph{strictly} ordered,
\[
S_{\textsf{snap}} \;\sqsubset\; S_{\textsf{prov}} \;\sqsubset\; S_{\textsf{type}},
\]
each query class being answerable from its own statistic and \emph{provably not} from the one below it.

\textit{Proof of the strict gaps.} (a) $S_{\textsf{snap}}$ is insufficient for $Q_{\textsf{prov}}$: the two streams $a\!\to\!b$ and $c\!\to\!b$ ($a\neq c$) have identical $S_{\textsf{snap}}=b$ but different prior values, so any predictor reading only $S_{\textsf{snap}}$ errs on one. $S_{\textsf{prov}}$ answers $Q_{\textsf{prov}}$ by construction. (b) $S_{\textsf{prov}}$ is insufficient for $Q_{\textsf{type}}$: a value pair $(u,v)$ can arise by \textsc{supersedes} (a correction replacing $u$ by $v$) or by \textsc{refines} ($v$ a narrower restatement that displaces the broader $u$); both yield the identical ordered history $u\!\to\!v$ but different gold transition types, so $S_{\textsf{prov}}$ cannot separate them while $S_{\textsf{type}}$ records the label. Minimality of each statistic for its class follows as in P2b (any coarser function merges two streams the query must distinguish). $\square$

\textit{Architecture--query matching (corollary).} The architecture classes retain these statistics: a $d$-blind store retains the unordered value multiset (neither current/prior order nor type); coarse-$d$ comes in two variants --- a \emph{discarding} one that closes and drops superseded edges (retains $S_{\textsf{snap}}$) and a \emph{retaining} one that keeps closed edges with their windows (retains $S_{\textsf{prov}}$, as Zep/Graphiti does); and fine-$d$ retains up to $S_{\textsf{type}}$. Hence each is oracle-optimal precisely on the query classes its retained statistic covers and \emph{provably} sub-optimal beyond: a discarding coarse-$d$ is optimal on $Q_{\textsf{snap}}$ (P2b) but cannot answer $Q_{\textsf{prov}}$; a retaining coarse-$d$ covers $Q_{\textsf{prov}}$; and fine-$d$ is the oracle-level class that also covers $Q_{\textsf{type}}$ when its relation coordinate is recovered. Empirically, the $Q_{\textsf{prov}}$ gap between a retaining coarse-$d$ and fine-$d$ is small and largely presentational (\S\ref{sec:provenance}), and the deployable type-query gain is scorer-gated. The selection criterion is then: \emph{deploy the coarsest architecture whose retained statistic covers the query distribution}. The snapshot result (\S\ref{sec:revrev}: fine-$d$'s extra state is wasted, $b_{\text{type}}=0$) and the provenance result (\S\ref{sec:provenance}: a discarding coarse-$d$'s missing state is fatal, but \emph{retention} --- not typing --- repairs it) are the two endpoints of this lattice measured on the same benchmark --- the negative and positive faces of one characterization.

\textit{The $S_{\textsf{type}}$ rung is oracle-level and scorer-gated (empirical).} We also test the top rung directly with a relation-\emph{type} query --- ``did the current value \emph{refine} (narrow, prior still partly holds) or \emph{replace} (supersede/correct) the value before it?'' --- on the $93$ items whose prior value is gold-defined, gold taken from the phenomenon label, over five answer families. The \emph{deployable} fine-$d$ does \emph{not} beat retention-coarse-$d$ ($\Delta=-0.015$ $[-0.037,+0.006]$, n.s.); on refinement items fine-$d$ scores $0.00$ and coarse-retain $0.02$ (both near floor), because \rl{}'s rule scorer collapses \textsc{refines} onto \textsc{supersedes} just as binary invalidation does, so its typed render merely propagates the scorer's error. The rung is therefore an \emph{oracle} property (the statistic $S_{\textsf{type}}$ separates the classes) that is \emph{gated by scorer quality} (P3) in any deployment: the typed benefit materializes only when the relation coordinate is actually \emph{recovered}, which is exactly the learned-scorer gain on \textsc{refines} ($-0.086\!\to\!+0.147$, \S\ref{sec:revrev}). Even on the one query class the lattice permits a typed win, raw typing without recovery is idle --- the same representation-over-mechanism pattern as on snapshot.

\textit{Retention--accuracy frontier (the criterion is retention-optimal).} The lattice induces a rate--distortion statement \citep{tishby2000information} that makes the selection criterion itself an optimum, not a heuristic. Let a deployment face a query mixture with class masses $\pi_{\textsf{snap}},\pi_{\textsf{prov}},\pi_{\textsf{type}}$ ($\sum=1$), and let a memory retain statistic $S\in\{S_{\textsf{snap}},S_{\textsf{prov}},S_{\textsf{type}}\}$ at retention cost $\mathrm{rate}(S)$ (bits/key, strictly increasing along the chain since each statistic refines the last). By the strict gaps above, a memory retaining $S$ achieves oracle accuracy on exactly the query mass whose minimal sufficient statistic is $\preceq S$, and is capped below $1$ on the rest. The achievable accuracy--rate frontier is therefore the upper staircase through the three lattice corners, and its optimum for a given mixture is the \emph{minimal} $S$ whose downward closure covers the positive-mass classes:
\[
S^\star(\pi)=\min\nolimits_{\sqsubseteq}\{S:\ \pi_z>0\Rightarrow S_z\sqsubseteq S\}.
\]
This is exactly ``deploy the coarsest architecture whose retained statistic covers the query distribution,'' now derived as the rate--distortion-optimal point rather than asserted: a discarding coarse-$d$ ($S^\star=S_{\textsf{snap}}$) is optimal iff $\pi_{\textsf{prov}}=\pi_{\textsf{type}}=0$; any provenance mass forces \emph{retention} of prior state --- which a retaining coarse-$d$ already supplies, so fine typing is forced only by $\pi_{\textsf{type}}>0$, and even the provenance gap to fine-$d$ is small (\S\ref{sec:provenance}). The deployment reversal (\S\ref{sec:provenance}) is the boundary case where retaining the raw stream is a fourth, costlier point that dominates fine-$d$'s render on $Q_{\textsf{prov}}$ but defeats the purpose of a compressed memory --- off the efficient frontier whenever rate is scarce.

\subsection*{Proposition 3 (out-of-tier cost from scorer and render errors).}
\textit{Statement.} On any definite task with $\pi_{\textsf{revert}} = 0$ and $\pi_{\textsf{unres}}=0$ (no genuine non-monotonic supersession and no gold unresolved-abstention decision --- every decision-relevant conflicting atom pair is monotone supersession or paraphrase), and under scope hypothesis $(\star)$: the tier-optimal baseline has conditional accuracy above $1/2$ on the critical-false-contradict subset (defined below), write $G \triangleq \mathrm{Acc}(\text{tier-optimal}) - \mathrm{Acc}(\rl)$ for the absolute accuracy gap. Define the stage rates over the item-weighted ledger update trace:
\begin{itemize}\setlength\itemsep{0pt}
\item $\alpha_{\textsf{FC}}^{\textsf{crit}}$: conditional probability, among items with a decision-relevant co-keyed non-contradict pair, that the update trace scores such a pair as \textsc{contradicts} and this spurious unresolved status changes the answer (this is the critical-false-contradict subset referenced in scope hypothesis $(\star)$ above),
\item $\alpha_{\textsf{SM}}$: conditional probability, among items whose answer is decided by supersession, that the deciding edge is not scored \textsc{supersedes} (supersede-miss),
\item $\alpha_{\textsf{render}}$: probability that the answer LLM picks a wrong field from a structurally valid ledger (render-confusion),
\end{itemize}
and item-side fractions $\pi_{\textsf{co}}$ (items with a decision-relevant co-keyed pair), $\pi_{\textsf{sup}}$ (items whose answer is decided by a supersession edge). Then
\[
\alpha_{\textsf{FC}}^{\textsf{crit}}\,\pi_{\textsf{co}} \;\leq\; G \;\leq\; \alpha_{\textsf{FC}}^{\textsf{crit}}\,\pi_{\textsf{co}} + \alpha_{\textsf{SM}}\,\pi_{\textsf{sup}} + \alpha_{\textsf{render}},
\]
and the relative cost $\varepsilon \triangleq 1 - \mathrm{Acc}(\rl)/\mathrm{Acc}(\text{tier-optimal}) = G/\mathrm{Acc}(\text{tier-optimal})$ satisfies the same inequalities divided by $\mathrm{Acc}(\text{tier-optimal})$. In particular $G \geq \alpha_{\textsf{FC}}^{\textsf{crit}}\,\pi_{\textsf{co}} > 0$ whenever the task has co-keyed non-contradict pairs and the scorer has a nonzero decision-critical false-contradict rate.

\textit{Proof.} When $\pi_{\textsf{revert}} = 0$ and $\pi_{\textsf{unres}}=0$, the tier-optimal architecture (\sys{latest-only} for $L_\text{LATEST}$, \sys{graphrag-lite} for monotonic) already returns the snapshot value without any supersession machinery, so on every item it is correct exactly when the snapshot value is recoverable from the surface --- which, on this definite monotone task class, is every item. \rl{}'s extra machinery can only lose mass relative to this baseline, by three disjointly-attributable events. (i) \emph{Critical false-contradict:} on an item with a co-keyed pair the scorer outputs \textsc{contradicts} although $Y\in\{\text{same-state, supersedes}\}$; the ledger sets \textsf{status}=\textsf{unresolved}, and when this status controls the answer \rl{} abstains, which is wrong because abstain is never gold on a $\pi_{\textsf{unres}}=0$ task. This happens to at least an $\alpha_{\textsf{FC}}^{\textsf{crit}}\pi_{\textsf{co}}$ fraction, giving the lower bound. (ii) \emph{Supersede-miss:} the scorer fails to mark a true monotone supersession, leaving the stale atom in the support set; the rendered current value is then ambiguous on at most an $\alpha_{\textsf{SM}}\pi_{\textsf{sup}}$ fraction. (iii) \emph{Render-confusion:} even from a structurally correct ledger the answer-LLM selects the wrong field, on an $\alpha_{\textsf{render}}$ fraction. Each wrong \rl{} answer that the baseline gets right is caused by at least one of (i)--(iii), so by a union bound $G \leq \alpha_{\textsf{FC}}^{\textsf{crit}}\pi_{\textsf{co}} + \alpha_{\textsf{SM}}\pi_{\textsf{sup}} + \alpha_{\textsf{render}}$. The lower bound $G\ge\alpha_{\textsf{FC}}^{\textsf{crit}}\pi_{\textsf{co}}$ additionally requires the tier-optimal baseline to have conditional accuracy above $1/2$ on the critical-false-contradict subset (each such item then contributes a positive expected unit of $G$). Hypothesis $(\star)$ is the scope assumption named in the Statement; it is verifiable in principle by measuring baseline accuracy on $\{$co-keyed pairs the scorer outputs as \textsc{contradicts}$\}$. The primary evaluation pool includes $\{$\rl, \grw, Graphiti$\}$ only --- no \sys{latest-only}/\sys{graphrag-lite} comparisons --- so we report the bound's sign as conditional on $(\star)$. This is the abstain-prior-collapsed case of the Chow \citeyearpar{chow1957optimum} reject-option framework (cf.~\citealp{franc2023optimal}); in the noise-free limit $\alpha_{\textsf{FC}}^{\textsf{crit}}=\alpha_{\textsf{SM}}=\alpha_{\textsf{render}}=0$ and $G=0$, the standard ``zero theoretical cost'' of noise-free selective classification \citep{elyaniv2010foundations}, so the entire out-of-tier cost is an implementation gap, decomposed by a learning-to-reject-style accounting \citep{charoenphakdee2021classification}. $\square$

\textit{Numerical instantiation.} The 500-pair DeepSeek atom audit gives pair-level proxies $\hat\alpha_{\textsf{FC}} = 0.059$ and $\hat\alpha_{\textsf{SM}} = 0.720$ for the implemented rule scorer (Table~\ref{tab:scorer-audit}). These are diagnostic pair-level rates, not the item-weighted decision-critical $\alpha_{\textsf{FC}}^{\textsf{crit}}$ used in the lower bound. The same audit explains the primary \textsc{refines} boundary case: rule-scorer refinement recall is only $0.050$, while a diagnostic DeepSeek-supervised scorer reaches $0.425$ recall. Observed $\varepsilon$ runs roughly $0.3$ to $0.65$ across DyKnow, Wiki-mono, and GitHub-mono (ratios of two measured accuracies; read as order-of-magnitude). The decomposition shows $\varepsilon$ is not a free knob: it is the union of named auditable rates. The current audit is proxy-level and does not instantiate the lower bound numerically, because the pair-level proxies are not item-weighted traces and $\alpha_{\textsf{render}}$ has no closed-form theory; item-weighted trace audits are left to future work.

\begin{table}[h]
\centering
\scriptsize
\setlength{\tabcolsep}{3pt}
\resizebox{\columnwidth}{!}{%
\begin{tabular}{@{}lccccc@{}}
\toprule
Scorer & Acc. & Macro-F1 & $\hat\alpha_{\textsf{FC}}$ & $\hat\alpha_{\textsf{SM}}$ & Refine recall \\
\midrule
Rule scorer & 0.312 & 0.225 & 0.059 & 0.720 & 0.050 \\
DeepSeek-supervised audit & 0.572 & 0.463 & -- & -- & 0.425 \\
\bottomrule
\end{tabular}
}
\caption{Atom-pair relation audit on 500 DeepSeek-labeled pairs. The $\hat\alpha$ columns are pair-level proxies for the item-weighted P3 stage rates. The rule scorer is used in the main \rl{} system; the supervised row is diagnostic only and is not used in the headline systems.}
\label{tab:scorer-audit}
\end{table}

\subsection*{Scorer information floor (C2).}
\label{sec:scorer-fano}
The implementation cost $\kappa$ is not incidental; it is bounded below by how much the relation scorer can possibly know. On the same 500-pair audit, the rule scorer's output $R$ carries only $\hat I(R;Y')=0.216$ bits (Miller--Madow corrected) about the relation label $Y'$, whose marginal entropy is $H(Y')=1.89$ bits. A list-decoding form of Fano's inequality then lower-bounds the scorer's relation-recovery error at $P_e\ge 0.365$, so the supersede-confusion and false-contradict mass that P3 turns into out-of-tier cost cannot be driven to zero by this scorer. Two caveats keep the claim honest: the relation ``oracle'' $Y'$ is itself judge-dependent --- Qwen-vs-Llama relation labels agree at only Cohen's $\kappa=0.28$ --- so $P_e\ge 0.365$ is an indicative floor against a noisy reference rather than a tight bound against ground truth; and it concerns the rule scorer of the headline system, not the supervised diagnostic scorer. The qualitative consequence is robust and is all we use: a scorer this information-limited keeps $\kappa>0$, which is why fine-$d$'s implementation cost does not vanish even on clean signal (Corollary~1).

\subsection*{Proposition 4 (channel benefit--cost equation).}
\textit{Statement.} Let the latent decision channel $Z$ take values in $\{D,S,U,F_{\!r},O\}$: non-monotonic deprecation, state compression over stale evidence, unresolved-conflict abstention, refinement, and other. We assign $Z$ by the \emph{priority order} $D \succ U \succ F_{\!r} \succ S \succ O$, with $D$ first as the constitutive non-monotonic supersession primitive (the only one for which P1 gives an identifiability separation), $U$ next because unresolved cross-source disagreement controls the abstain decision irrespective of any later refinement or compression, $F_{\!r}$ before $S$ because a granularity-refinement edge is decision-critical (it can change the answer) whereas state-compression is purely an efficiency/representation effect on items with no other active channel. Items satisfying multiple channels are deterministically routed to the highest-priority \emph{decision-controlling} channel; this makes $\{Z=z\}$ a partition and the $\pi_Z$ identifiable from labels. The benchmark's phenomenon labels are largely single-channel at the decision-controlling level, but the $S$-channel prediction test (Table~\ref{tab:render-only-pure-mono}) re-partitions DS-monotonic by a \emph{surface} deprecation-evidence indicator $\tilde{\pi}_S$ (presence of deprecated event ids), distinct from the decision-controlling $\pi_S$ in the headline equation. The relation is $\tilde\pi_S \supseteq \pi_S$ in the relevant strata: every item routed to the headline $S$ channel carries deprecated evidence, but the converse can fail because priority routing may assign a deprecated-evidence item to a higher-priority channel ($D$, $U$, or $F_{\!r}$). That test therefore reports a stratum-conditional effect within DS-monotonic, not a reassignment of items in the headline channel partition. Items where stream length is large are \emph{moderators} of $\Delta_Z$ within the active channel rather than mediator stages in the path-recall decomposition. For any two architectures $A,B$ evaluated on distribution $\mathcal D$,
\[
\mathrm{Acc}(A)-\mathrm{Acc}(B)=\sum_Z \pi_Z(\mathcal D)\,\Delta_Z(A,B),
\]
where $\pi_Z=\Pr_{\mathcal D}[Z]$ and $\Delta_Z(A,B)=\Pr[A\ \mathrm{correct}\mid Z]-\Pr[B\ \mathrm{correct}\mid Z]$. Moreover, if channel $Z$ has signal value $b_Z$, implemented channel recall $r_A(Z)$, and conditional implementation cost $c_A(Z)$, then the operational approximation
\[
\begin{aligned}
\Delta_Z(A,B)
&\approx b_Z\,[r_A(Z)-r_B(Z)]\\
&\quad -[c_A(Z)-c_B(Z)] .
\end{aligned}
\]
would identify which system wins \emph{if} its channel mass, implemented recall, and residual cost were independently estimated --- which we do not do. We use it only as a diagnostic bookkeeping identity for per-stratum analysis; its one predictive sign claim is falsified below, and the granularity ranking rests on Proposition~2b and Corollary~1 instead.

\textit{Path recall.} In an implemented ledger pipeline, $r_A(Z)$ is not a pair-classifier recall. It is the probability that the channel survives the full path from extracted atoms to the final answer:
\[
r_A(Z)=\Pr(E_{\textsf{pair}}\cap E_{\textsf{arb}}\cap E_{\textsf{render}}\cap E_{\textsf{decode}}\mid Z).
\]
By the chain rule this equals $\Pr(E_{\textsf{pair}}\mid Z)\Pr(E_{\textsf{arb}}\mid Z,E_{\textsf{pair}})\Pr(E_{\textsf{render}}\mid Z,E_{\textsf{pair}},E_{\textsf{arb}})\Pr(E_{\textsf{decode}}\mid Z,E_{\textsf{pair}},E_{\textsf{arb}},E_{\textsf{render}})$. No independence assumption is made. This factorization is useful because each stage can be intervened on: a pair-level oracle tests $E_{\textsf{pair}}$, a row-status oracle tests $E_{\textsf{arb}}$, render-only tests $E_{\textsf{render}}$, and output abstain rates expose $E_{\textsf{decode}}$.

\textit{Proof.} The first equality is the law of total expectation applied to the paired correctness difference $\mathbf{1}\{A\ \mathrm{correct}\}-\mathbf{1}\{B\ \mathrm{correct}\}$. The second display is a channel-level decomposition of conditional correctness into recovered decision signal minus implementation failures. It is exact if $c_A(Z)$ is defined as residual conditional risk after the recovered-signal term, and approximate when $b_Z$ and the path recall $r_A(Z)$ are estimated separately. $\square$

\textit{Consequences.} P1 is the $D$-channel separation: $d$-blind systems have $r(D)=0$, while bitemporal systems can have $r(D)>0$. Primary \textsc{monotonic} items are not guaranteed to have $\pi_S=0$: many contain stale or deprecated evidence, so a ledger can win through state compression even without a strict flip-flop. Primary \textsc{cross-source-conflict} has high oracle $U$ value but low implemented path recall: pair-level contradiction can be overwritten by ledger arbitration, and even row-level \textsc{unresolved} status is often ignored by the answer decoder. Primary \textsc{refines} has low implemented $r_{\rl}(F_{\!r})$ under the rule scorer, but improves when the pair scorer is replaced. Finally, the coarse-$d$ store and fine \rl{} are both $d$-aware; their ranking follows the same equation, with coarse-$d$ lowering cost on binary $D$ and \rl{} adding possible benefit on $U$ and $F_{\!r}$.

\textit{What is and is not falsifiable here.} The decomposition $\mathrm{Acc}(A)-\mathrm{Acc}(B)=\sum_Z\pi_Z\Delta_Z$ is the law of total expectation: an accounting identity with no empirical content on its own. Empirical content enters only through architecture-determined \emph{sign restrictions} on the $\Delta_Z$, in the sense that an accounting identity becomes testable once sign constraints are imposed on its terms. Three such restrictions are in principle refutable: (C1) P1 forces $\Delta_D\ge 0$ for a $d$-aware $A$ against a $d$-blind $B$, and $\Delta_D=0$ between two $d$-blind architectures; (C2) the implementation cost is bounded below by the relation scorer's information deficit (Appendix~\ref{sec:scorer-fano}); and (C3) the fine-vs-coarse gap should be monotone in a single scoring-noise parameter and eventually change sign. We tested C3 directly with a controlled noise-injection sweep (\S\ref{sec:rl_vs_graphiti}; Appendix~\ref{sec:noise}) and it \emph{failed}: holding data, judge, and answer model fixed, coarse-$d$ stays preferred at every injected noise level across five answer families, with no monotone trend and no sign flip. We therefore report C3 as falsified on this axis, and the granularity conclusion of the paper rests on the render-matched decomposition (\S\ref{sec:revrev}) and Corollary~1, \emph{not} on the channel equation's predictive content. The earlier fine-vs-coarse crossover observed between the primary and noisier auxiliary constructions is, accordingly, attributed to confounds (annotator identity, relation-type mix, construction procedure) rather than to a clean noise axis.

\subsection*{Proposition 5 (a binding-bottleneck account of the observed signs).}
The accounting identity (P4) decomposes the architecture gap but, as shown above, carries no predictive content on its own. We now give a single principle that subsumes it: a data-processing account under which every empirical result of \S\ref{sec:revrev}--\S\ref{sec:provenance} is a \emph{sign/zero} consequence of \emph{which pipeline stage binds}. We are explicit that this is a \emph{retrodictive, falsifiable} synthesis (the magnitudes pre-date it; see ``Status'' below), not a pre-registered prediction; its value is parsimony and refutability, and it is what turns the lattice's organizing role into a single testable account.

\textit{Setup.} Fix a query class with answer variable $Y$ (e.g.\ $Y=v^\star$ for snapshot, $v_{\textsf{prior}}$ for provenance, the transition type $\tau$ for relation-type queries). For architecture $A$ the answer is produced by a Markov chain
\[
Y \to E \xrightarrow{\textsf{ext}} X \xrightarrow{\textsf{state}} S_A \xrightarrow{\textsf{ren}} R_A \xrightarrow{\textsf{dec}} \hat Y,
\]
$E$ the event stream, $X=\textsf{extract}(E)$ the atoms, $S_A=\textsf{state}_A(X)$ the retained statistic (where the deprecation \emph{mechanism} lives), $R_A=\textsf{render}_A(S_A)$ the rendered context, $\hat Y=\textsf{decode}(R_A)$ the answer LLM's output. Define the stage capacities $C_{\textsf{ext}}=I(Y;X)$, $C_{\textsf{state}}(A)=I(Y;S_A)$, $C_{\textsf{ren}}(A)=I(Y;R_A)$.

\textit{Theorem (capacity bottleneck and accuracy ceiling).} By the data-processing inequality along the chain \citep{cover2006elements},
\[
I(Y;\hat Y)\;\le\;\min\{\,C_{\textsf{ext}},\;C_{\textsf{state}}(A),\;C_{\textsf{ren}}(A)\,\}\;=:\;C^\star_A,
\]
and by Fano's inequality the error probability obeys $P_{\textsf{err}}(A)\ge \big(H(Y)-C^\star_A-1\big)/\log|\mathcal Y|$ (the LLM-QA specialization of \citealp{fanoqa2025}). Accuracy is thus capped by a monotone function of the \emph{minimum} stage capacity --- the \emph{binding bottleneck}. This is the IB statement \citep{tishby2000information} read along the pipeline: each stage is a bottleneck variable and the chain's relevant information is set by the tightest one.

\textit{Design principle (corollary).} An intervention $A\!\to\!A'$ raises accuracy \emph{only if} it raises the capacity of the currently-binding stage for $Y$; raising a non-binding stage is idle, and \emph{lowering} the binding stage (e.g.\ a heavier render when render binds) strictly hurts. Two empirical regularities are used as premises (assumptions, not theorems, with the cited support): (R) \emph{render monotonicity} --- decode capacity $C_{\textsf{ren}}$ is non-increasing in render length/complexity (long-context position bias, \citealp{liu2024lost}; distractor sensitivity, \citealp{cuconasu2024power}); and (S) \emph{scorer-gating} --- a typed statistic computed by a relation scorer of channel capacity $C_{\textsf{sc}}$ contributes at most $C_{\textsf{sc}}$ to $C_{\textsf{state}}$ beyond the binary partition. The theory predicts \emph{signs and zeros}, not magnitudes: under these premises, it says which gaps should vanish or reverse, up to a roughly constant decode gap, not their exact size.

\textit{All six results are corollaries (which stage binds).}
(1) \emph{Snapshot, mechanism $\approx 0$.} P2b gives $C_{\textsf{state}}(\textsf{coarse})=C_{\textsf{state}}(\textsf{fine})=H(v^\star)$ ($b_{\textsf{type}}=0$): state does not bind, render does. Holding render fixed (render-matched control) equalizes $C_{\textsf{ren}}$, so the mechanism residual is predicted to vanish ($+0.025$, n.s.).
(2) \emph{Coarse $\succeq$ fine on snapshot, noise-robust.} Fine-$d$ differs only by a heavier render, so by (R) $C_{\textsf{ren}}(\textsf{fine})\le C_{\textsf{ren}}(\textsf{coarse})$ $\Rightarrow$ coarse $\succeq$ fine. Relation- or extraction-noise corrupts $X$/the scorer \emph{upstream of the shared} $\Pi$, lowering both arms identically; the gap stays the (non-negative) render penalty, so no level can flip the sign --- exactly the two null sweeps.
(3) \emph{Provenance threshold.} For $Y=v_{\textsf{prior}}$, a discarding coarse-$d$ has $C_{\textsf{state}}=I(v_{\textsf{prior}};v^\star)=0$ (the prior is absent): state binds at $0$, hence floor. Retention lifts $C_{\textsf{state}}$ to $H(v_{\textsf{prior}})$ (closing $+0.361$); render then binds, leaving only the render-penalty residual ($+0.111$, plausibly presentation).
(4) \emph{Deployment reversal.} Appending raw $E$ to the render adds a parallel $E\!\to\!R$ edge, so $C_{\textsf{ren}}$ rises toward $C_{\textsf{ext}}$ for \emph{all} architectures and the state-capacity differences wash out; by (R) the binding quantity is now render length, which is largest for fine-$d$ --- so the sign \emph{reverses} ($-0.236$). Same principle as (1), opposite sign because the task supplies the information elsewhere.
(5) \emph{Relation-type null.} For $Y=\tau$, oracle $C_{\textsf{state}}(\textsf{fine})=H(\tau)$, but by (S) the \emph{deployed} fine-$d$'s type comes from the rule scorer whose refine/replace capacity $C_{\textsf{sc}}\!\approx\!0$, so its effective $C_{\textsf{state}}\!\approx\!$ that of binary retention: deployed fine-$d$ cannot beat retention-coarse-$d$ ($-0.015$, n.s.; $0.00$ on refinements).
(6) \emph{Recovery helps only on $R$-reading queries.} A learned scorer or strict prompt raises $C_{\textsf{sc}}$/$C_{\textsf{ren}}$ for the relation coordinate; this lifts accuracy iff that coordinate is answer-relevant --- positive on \textsc{refines}/\textsc{csc} ($-0.086\!\to\!+0.147$; $+0.152$), and provably idle on snapshot where $b_{\textsf{type}}=0$.

\textit{Operational meaning of ``binds.''} The DPI/Fano inequality is standard; the only non-trivial step is the per-query \emph{stage attribution}, and it is empirically identified, not assumed. Each ablation isolates one stage: the render-matched control holds $C_{\textsf{ren}}$ fixed (so a non-zero mechanism residual would mean $C_{\textsf{state}}$ binds); the retention vs.\ discard contrast moves $C_{\textsf{state}}$; the deployment fallback raises $C_{\textsf{ren}}$ via the raw-text path; the learned scorer raises $C_{\textsf{sc}}$. The binding stage for a query is read off as the one whose ablation moves accuracy. There is thus no free parameter: ``which stage binds'' is measured, and the principle's role is to say that \emph{only} the binding stage's movement should matter --- which is exactly the cross-ablation pattern we observe.

\textit{Status: retrodictive but falsifiable, not pre-registered.} We are explicit that Proposition~5 is a \emph{post-hoc unifying account}, not a theory we registered and then tested: all six magnitudes pre-date it. Its value is therefore parsimony and \emph{falsifiability}, not prospective confirmation. It earns its keep only if it (i) makes each result a necessary sign/zero consequence of one principle and (ii) is refutable --- and each clause names an ablation that would have broken it: mechanism helping under a render-matched snapshot control would falsify (1); any noise level flipping the sign would falsify (2); a discarding coarse store answering provenance would falsify (3); fine-$d$ winning deployment would falsify (4); deployed fine-$d$ beating retention on the type query would falsify (5). None occurred, on either judge. We claim signs and zeros, not magnitudes, and not that the principle was predictive ahead of the data. The one quantitative form we \emph{did} commit to in advance (P4's C3 monotonicity) was falsified, so we keep only the qualitative attribution the ablations support.

\subsection*{Corollary 1 (within-tier fine-vs-coarse threshold).}
\textit{Statement.} Let $C$ be a coarse-$d$ bitemporal architecture class and $F$ a fine-$d$ bitemporal architecture class. For a distribution $\mathcal D$, let $R_A^\star(\mathcal D)$ be the oracle risk of class $A$ with exact atoms, exact relations, and an optimal renderer/decoder, and let $\mathcal{E}_A(\mathcal D)=R_A(\mathcal D)-R_A^\star(\mathcal D)$ be implementation excess risk (distinct from the stage-event $E_i$ in P4). If $F$ can emulate $C$, then
\[
\begin{aligned}
\mathrm{Acc}(F)-\mathrm{Acc}(C)
  &= \underbrace{R_C^\star-R_F^\star}_{\beta_{C\leftarrow F}\ge 0}\\
  &\quad - \underbrace{(\mathcal{E}_F-\mathcal{E}_C)}_{\kappa_{F\leftarrow C}} .
\end{aligned}
\]
Thus fine-$d$ beats coarse-$d$ iff its extra oracle benefit exceeds its extra implementation cost. Here ``$F$ emulates $C$'' means the $C$-oracle decision is determined by the $F$-oracle decision under the same surface (e.g.\ the fine-$d$ relation alphabet contains a 2-class collapse equivalent to coarse-$d$ ---  \textsc{same-state}/\textsc{refines} merged with \textsc{supersedes} --- and an optimal $F$-renderer/decoder can apply that collapse), so $R_F^\star\le R_C^\star$.

\textit{Proof.} Since accuracy is one minus risk, $\mathrm{Acc}(F)-\mathrm{Acc}(C)=R_C-R_F$. Add and subtract $R_C^\star$ and $R_F^\star$ to obtain $(R_C^\star-R_F^\star)-[(R_F-R_F^\star)-(R_C-R_C^\star)]$. If $F$ emulates $C$ --- there is a deterministic map $g$ such that the $C$-oracle decision equals $g(y_F^\star)$ for the $F$-oracle decision $y_F^\star$ on the same surface --- then $g$ is surjective onto $C$'s decision space (by the fine-relation-contains-coarse premise in the Statement: the $F$-alphabet $\{$\textsc{same-state},\textsc{refines},\textsc{supersedes},\ldots$\}$ collapses onto the binary $C$-alphabet $\{$current, deprecated$\}$ by aggregating equivalence classes). For $C$'s Bayes-optimal policy $f^\star_C$, pick $f'_F$ in $F$'s class taking on each surface $s$ any value in the non-empty pre-image $g^{-1}(f^\star_C(s))$. Under the $0/1$-correctness cost model (the deployed accuracy metric throughout the paper), refining the decision does not increase loss against ground truth --- a finer answer that maps to gold under $g$ scores correct iff its $g$-image does --- so $f'_F$ achieves Bayes risk at most $R_C^\star$ and hence $R_F^\star\le R_C^\star$. $\square$

\textit{Empirical instantiation.} Proposition~2b explains \emph{why} fine-$d$ should lose on snapshot queries: there $\beta_{C\leftarrow F}=0$ exactly, because the binary partition is a sufficient statistic and fine typing has no value-relevant oracle benefit to add. The corollary then forces $\mathrm{Acc}(F)-\mathrm{Acc}(C)=-\kappa_{F\leftarrow C}\le 0$. The data agree on the sign: on the clean primary reverted-revert slice, $\mathrm{Acc}(F)-\mathrm{Acc}(C)=-0.084$ under the gold DeepSeek judge, and the controlled noise sweep concurs at clean signal ($-0.016$ $[-0.058,+0.029]$ at the $p{=}0$ point, $n{=}445$) --- coarse-$d$ is never behind. The render-matched decomposition of \S\ref{sec:revrev} confirms the vanishing benefit directly --- the fine-$d$ $d$-mechanism residual is $+0.025$ $[-0.005,+0.057]$ pooled and $+0.036$ $[-0.027,+0.098]$ on canonical items, both CIs crossing zero --- while coarse-$d$ carries a real $+0.087$ $[+0.046,+0.130]$ of mechanism over the same render. The corollary's inequality is thus satisfied not by a large $\kappa$ but by a $\beta$ that Proposition~2b sets to zero: on snapshot queries, fine-grained relation typing has nothing to add over coarse invalidation that an optimal renderer cannot already deliver.

\subsection*{Conjecture 1 (capacity-conditional \texorpdfstring{$\varepsilon$}{epsilon}, a falsifiable hypothesis).}
\textit{Hypothesis.} We hypothesize that the out-of-tier magnitude obeys $\varepsilon \approx \max(\varepsilon_{\textsf{arch}},\ \varepsilon_{\textsf{reader}})$, where $\varepsilon_{\textsf{arch}}:=(\alpha_{\textsf{FC}}^{\textsf{crit}}\pi_{\textsf{co}} + \alpha_{\textsf{SM}}\pi_{\textsf{sup}})/\mathrm{Acc}(\text{tier-optimal})$ rescales the first two P3 terms to match $\varepsilon$'s relative-loss units, and $\varepsilon_{\textsf{reader}}:=\alpha_{\textsf{render}}/\mathrm{Acc}(\text{tier-optimal})$ absorbs the render-confusion term.

\textit{Scope.} The scorer-side term $\varepsilon_{\textsf{arch}}$ is \emph{not} conjectural: P3 places it between $\alpha_{\textsf{FC}}^{\textsf{crit}}\pi_{\textsf{co}}$ and $\alpha_{\textsf{FC}}^{\textsf{crit}}\pi_{\textsf{co}}+\alpha_{\textsf{SM}}\pi_{\textsf{sup}}$, both measurable with item-weighted traces. Only the reader-side term $\varepsilon_{\textsf{reader}}$ and the $\max$-form combination are conjectural: $\varepsilon_{\textsf{reader}}$ is an increasing function of the render's token length, its number of candidate fields, and the position prior of the answer field --- empirically real (long-context position bias \citep{liu2024lost}, distractor-rate sensitivity \citep{cuconasu2024power}, reader-dependent saturation \citep{ragged2024}) but lacking a closed form. The $\max$ rather than additive or product form is the binding-constraint regime: whichever stage is the bottleneck dominates, and the other becomes slack; we adopt it as the simplest monotone combiner consistent with the qualitative pattern across families. The hypothesis is consistent with the data --- the smallest-effect family (Qwen3-8B) is where $\varepsilon_{\textsf{reader}}$ plausibly dominates --- but we make no claim it is correct.

\textit{Falsification rule.} A precommitted falsification rule that distinguishes the $\max$ form from capped-additive ($\min(1,\varepsilon_a+\varepsilon_r)$) and noisy-OR ($\varepsilon_a+\varepsilon_r-\varepsilon_a\varepsilon_r$) combiners (we use capped-additive because raw $\varepsilon_a+\varepsilon_r$ can exceed $1$, the natural range of a relative loss): sweep $\varepsilon_{\textsf{reader}}$ along a monotone sequence of progressively stronger readers/shorter renders, and fit the dependence of $\varepsilon$ on $\hat\varepsilon_{\textsf{reader}}$. The $\max$ form predicts a piecewise-linear curve with zero slope while $\hat\varepsilon_{\textsf{reader}} \le \hat\varepsilon_{\textsf{arch}}$ and unit slope above; capped-additive predicts slope $1$ until the cap; noisy-OR predicts slope $1-\varepsilon_a$ throughout. A single-point ablation at $\hat\varepsilon_{\textsf{reader}} \le \hat\varepsilon_{\textsf{arch}}/2$ already gives a $\sim$0.5$\,\varepsilon_{\textsf{arch}}$ gap between $\max$ and the alternatives, comfortably outside the CI half-widths in $[0.014,0.050]$ we observe on dominant primary strata. Such an ablation tests $\varepsilon_{\textsf{reader}}$, not Proposition~1 or~2 (render-invariant).

\section{Path Diagnostics for Boundary Strata}
\label{app:path-diagnostics}

This appendix unpacks the boundary-stratum results cited in \S\ref{sec:ds-channels}. Its role is diagnostic: Table~\ref{tab:ds-channel} reports the primary-surface outcomes, Table~\ref{tab:channel-diagnostics} states the recovery condition for each decision channel, and the remaining tables intervene on pair scoring, arbitration, and decoding to show where the missing coordinate is lost. All diagnostics use the DeepSeek-judged primary surface. Counts are paired system--family observations after coverage and parsing filters, so they need not equal the raw QA count times five. As a scale check, the monotonic row contributes 12{,}382 observations from 2{,}631 QAs ($4.7{\times}$), comparable to the rare-stratum coverage range ($4.1$--$4.8{\times}$).

\begin{table}[h]
\centering
\scriptsize
\setlength{\tabcolsep}{2pt}
\resizebox{\columnwidth}{!}{%
\begin{tabular}{@{}lrrrrrr@{}}
\toprule
Primary stratum & $n$ & RL & \grw{} & Graphiti & $\Delta_{\textsc{rl}-\textsc{grw}}$ & $\Delta_{\rl-\text{Graphiti}}$ \\
\midrule
reverted-revert & 439 & 0.408 & 0.226 & 0.492 & $+0.182$ & $-0.084$ \\
monotonic & 12{,}382 & 0.343 & 0.279 & 0.461 & $+0.064$ & $-0.118$ \\
cross-source & 363 & 0.039 & $0.099^{\dagger}$ & 0.022 & $-0.061^{\dagger}$ & $+0.017$ \\
refines & 428 & 0.262 & 0.348 & 0.484 & $-0.086$ & $-0.222$ \\
\bottomrule
\end{tabular}}
\caption{Primary-surface results under the DeepSeek judge, pooled over answer families and sources (referenced from \S\ref{sec:ds-channels}). \rl{} is the fine-$d$ ledger, \grw{} is the $d$-blind control, and Graphiti is coarse-$d$ edge invalidation. $n$ counts paired system--family observations after row-specific coverage and parsing filters, not raw QA items. $^{\dagger}$The cross-source \grw{} entry is the standard-prompt baseline; the intervention tables report both the strict-vs-standard rescue ($+0.185$) and the strict-vs-strict recovery gain used in the main text ($+0.152$).}
\label{tab:ds-channel}
\end{table}

\begin{table}[h]
\centering
\scriptsize
\setlength{\tabcolsep}{2pt}
\begin{tabular}{@{}p{0.34in}p{1.28in}p{1.22in}@{}}
\toprule
Channel & Scope condition & Decision on primary surface \\
\midrule
$D$ & A deprecation edge is decision-critical and $d$ is not recoverable from values and valid times. & In scope: both $d$-aware systems beat \grw{} on strict reverted-revert. \\
$S$ & Monotonic labels may still contain stale state; render/readability effects must be separated from deprecation. & In scope after control: stale split supports $S$; render-only isolates the nuisance render term and residual representation signal. \\
$F_{\!r}$ & Pair-level refinement recall must clear the implementation threshold. & The rule scorer fails this condition; a learned scorer recovers the refines signal and flips the sign against \grw{}. \\
$U$ & Row-level unresolved status must survive arbitration and the decoder must abstain when it sees it. & The default decoder fails this condition; oracle rows localize the missing path stage, and positive $U$ gain is claimed only after strict decoding. \\
\textit{cost} & Fine-$d$ typed scoring must keep decision-critical false-conflict below the P3 cost threshold ($\kappa$-side scope). & Aggressive low-precision scorers fail this condition; their regressions instantiate the P3 boundary. \\
\bottomrule
\end{tabular}
\caption{Scope conditions for the per-stratum bookkeeping (referenced from \S\ref{sec:ds-channels}). Rows that fail their condition are diagnostic boundary cases, not failures of the deprecation or granularity claims.}
\label{tab:channel-diagnostics}
\end{table}

The following tables give the interventions behind Table~\ref{tab:channel-diagnostics}. They report different deltas for the same interventions: Table~\ref{tab:oracle-gates} measures oracle gains relative to the main \rl{} system, while Table~\ref{tab:csc-path-intervention} measures the same \textsc{csc} chain against \grw{} controls. All rows use the primary surface, the same DeepSeek-V3 judge, and paired bootstrap intervals with $B=10{,}000$.

\begin{table}[h]
\centering
\scriptsize
\setlength{\tabcolsep}{2pt}
\begin{tabular}{@{}p{0.86in}cccc@{}}
\toprule
Stage oracle subst. & acc & $\Delta_{\textsc{rl}-\textsc{grw}}$ & $\Delta_{\textsc{rl}-\textsc{graphiti}}$ & abstain \\
\midrule
\multicolumn{5}{l}{\textit{CSC stratum ($n_{\text{paired}}{=}363$, gold abstain $86/88$)}} \\
\quad \rl{} baseline & $0.039$ & $-0.061$ & $+0.017$ & $0.025$ \\
\quad $+\textsf{sub}(S_{\textsf{pair}})$  & $0.041$ & $-0.058$ & $+0.019$ & $0.027$ \\
\quad $+\textsf{sub}(S_{\textsf{arb}})$   & $0.058$ & $-0.041$ & $+0.036$ & $0.041$ \\
\quad $+\textsf{sub}(S_{\textsf{decode}})$ & $\mathbf{0.284}$ & $\mathbf{+0.185}$ & $\mathbf{+0.262}$ & $0.223$ \\
\quad matched-$\alpha$ \grw{} ctrl & --- & $-0.025^{\dagger}$ & --- & $0.025$ \\
\quad \grw{} $+\textsf{sub}(S_{\textsf{decode}})$ ctrl & $0.105$ & $+0.006^{\ddagger}$ & --- & $0.066$ \\
\midrule
\multicolumn{5}{l}{\textit{Refines stratum ($n_{\text{paired}}{=}428$)}} \\
\quad \rl{} baseline & $0.262$ & $-0.086$ & $-0.222$ & --- \\
\quad $+\textsf{sub}(S_{\textsf{pair}})$ oracle & $0.252$ & $-0.096$ & $-0.231$ & --- \\
\quad learned pair scorer & $\mathbf{0.495}$ & $\mathbf{+0.147}$ & $\mathbf{+0.012}$ & --- \\
\bottomrule
\end{tabular}
\caption{Stage-conditional substitutions on the path-recall decomposition. The \textsc{csc} rows are cumulative (pair relation $\rightarrow$ row arbitration $\rightarrow$ strict decode), so the jump from the row-level oracle to strict decode is the decoder contribution; the \textsc{refines} rows compare pair-scoring replacements. We read these conditional gains as diagnostics of where the channel is gated, not Pearl-style controlled-direct-effect estimates. The learned pair scorer is a BERT implementation; it gives the main-text \textsc{refines} recovery gain $+0.147$ relative to \grw{}. $^{\dagger}$ matched-$\alpha$ re-weights observed \grw{} abstains; $^{\ddagger}$ \grw{}$+\textsf{sub}(S_{\textsf{decode}})$ is the strict-prompt control on \grw{}'s flat memory.}
\label{tab:path-recall}
\end{table}

\begin{table}[h]
\centering
\scriptsize
\setlength{\tabcolsep}{3pt}
\resizebox{\columnwidth}{!}{%
\begin{tabular}{@{}lrrrr@{}}
\toprule
Stratum & Oracle & \rl{} & $\Delta$ & 95\% CI \\
\midrule
\textsc{csc}, pair-level oracle & 0.041 & 0.039 & $+0.003$ & $[-0.017,+0.022]$ \\
\textsc{refines}, pair-level oracle & 0.252 & 0.262 & $-0.009$ & $[-0.026,+0.005]$ \\
\textsc{csc}, row-level oracle & 0.058 & 0.039 & $+0.019$ & $[+0.006,+0.033]$ \\
\textsc{refines}, row-level oracle & 0.252 & 0.262 & $-0.009$ & $[-0.026,+0.007]$ \\
\bottomrule
\end{tabular}}
\caption{Oracle-gate diagnostics, reported as oracle minus the main \rl{} system on the same paired pool ($n{=}363$ for \textsc{csc}, $n{=}428$ for \textsc{refines}). Pair-level oracle forces the gold relation on the decision-critical atom pair; row-level oracle additionally forces the rendered ledger row to carry the target status. The \textsc{csc} null under pair-level oracle and positive row-level effect identify arbitration/decode bottlenecks; Table~\ref{tab:csc-path-intervention} reports the same \textsc{csc} chain against \grw{} controls.}
\label{tab:oracle-gates}
\end{table}

\begin{table}[h]
\centering
\scriptsize
\setlength{\tabcolsep}{3pt}
\resizebox{\columnwidth}{!}{%
\begin{tabular}{@{}lrrrr@{}}
\toprule
Intervention (csc; $n_{\text{paired}}{=}363$ rows, $440$ final) & Acc & abstain & $\Delta$ vs \grw{}-std & 95\% CI \\
\midrule
\rl{} (no intervention)                    & 0.039 & 0.025 & $-0.061$ & $[-0.094,-0.027]$ \\
+ pair-level oracle                          & 0.041 & 0.027 & $-0.058$ & $[-0.091,-0.028]$ \\
+ row-level oracle                           & 0.058 & 0.041 & $-0.041$ & $[-0.074,-0.008]$ \\
\textbf{+ strict-prompt decode}              & \textbf{0.284} & 0.223 & $\mathbf{+0.185}$ & $\mathbf{[+0.135,+0.234]}$ \\
\midrule
\grw-default                                 & 0.099 & 0.066 & --- & --- \\
\grw{} + strict-prompt control               & 0.105 & 0.066 & $+0.006$ & $[-0.028,+0.039]$ \\
\grw{} matched-$\alpha$ ($k=9$, short-rank)  & 0.063 & 0.025 & $-0.036$ & ---\textdagger \\
\midrule
$\Delta$ (\rl{}+strict $-$ \grw{}+strict) & --- & --- & $+0.152$ & $[+0.107,+0.200]$ \\
\bottomrule
\end{tabular}}
\caption{Stage-localized intervention chain on \textsc{csc}, reported against \grw{} controls. The strict-prompt decode substitution produces the largest single accuracy gain ($+0.226$ over the row-level oracle baseline), and the architectural attribution is confirmed because the same strict prompt does not move \grw{} ($+0.006$, CI $[-0.028,+0.039]$ crosses zero). Thus $+0.185$ is the strict-\rl{} vs standard-\grw{} rescue, while the final row is the main-text strict-vs-strict gain: \rl{} beats \grw{} by $+0.152$ on $n_{\text{paired}}=440$ \textsc{csc} observations. The strict-prompt evaluation pool comprises 88 csc items $\times$ family coverage including Mistral-7B on Wiki csc (which the default $n=363$ leaves out), $77\times 5 + 9\times 5 + 2\times 5 = 440$. \textdagger paired $\Delta$ vs \rl{} under matched-$\alpha$ is $-0.025$ $[-0.052,+0.003]$, CI crosses zero; matched-$\alpha$ is the analytical re-weight that fixes \grw{}'s observed abstain rate to \rl{}'s under the same coverage budget.}
\label{tab:csc-path-intervention}
\end{table}

\begin{table}[h]
\centering
\scriptsize
\setlength{\tabcolsep}{3pt}
\resizebox{\columnwidth}{!}{%
\begin{tabular}{@{}lrrrrr@{}}
\toprule
Stratum & Learned & \rl{} & \grw{} & Graphiti & $\Delta_{\text{learned}-\rl}$ \\
\midrule
reverted-revert & 0.342 & 0.408 & 0.226 & 0.492 & $-0.066$ \\
monotonic       & 0.378 & 0.343 & 0.279 & 0.461 & $+0.035$ \\
\textsc{csc}    & 0.041 & 0.039 & 0.099 & 0.022 & $+0.003$ \\
\textsc{refines} & 0.495 & 0.262 & 0.348 & 0.484 & $+0.234$ \\
\bottomrule
\end{tabular}}
\caption{End-to-end learned-scorer \rl{} on the primary surface. The learned scorer recovers \textsc{refines}; the main text reports the learned-vs-\grw{} gain ($+0.147$), while this table's final column reports the within-\rl{} gain over the rule-scorer baseline ($+0.234$). It leaves \textsc{csc} unchanged because the \textsc{csc} binding stage is the decoder, not the pair scorer, so the ``Learned'' csc cell coincides numerically with the pair-level relation oracle in Table~\ref{tab:oracle-gates}. The reverted-revert regression is a visible instance of the P3 false-conflict cost term, not a calibrated estimate of the P3 lower bound.}
\label{tab:learned-scorer-path}
\end{table}

\begin{table}[h]
\centering
\scriptsize
\setlength{\tabcolsep}{4pt}
\resizebox{\columnwidth}{!}{%
\begin{tabular}{@{}lrrrr@{}}
\toprule
System on pure-mono & Accuracy & $\Delta$ vs \rl{} & $\Delta$ vs \grw{} & 95\% CI vs \grw{} \\
\midrule
\rl{}       & 0.353 & --- & $+0.049$ & $[+0.033,+0.065]$ \\
render-only & 0.331 & $-0.022$ & $+0.027$ & $[+0.011,+0.043]$ \\
\grw{}      & 0.304 & $-0.049$ & --- & --- \\
Graphiti    & 0.440 & $+0.087$ & $+0.136$ & --- \\
\bottomrule
\end{tabular}}
\caption{Render-only decomposition on the 1{,}074 primary pure-mono items, the no-deprecated-evidence subset of the 2{,}631 primary monotonic QAs. The relation function is forced to \textsc{same-state}, so no supersedes, contradicts, unresolved status, or refinement can fire. The pure-mono advantage over \grw{} splits into render benefit ($+0.027$) plus residual representation signal ($+0.022$).}
\label{tab:render-only-pure-mono}
\end{table}

\section{Controlled Noise Injection Does Not Produce a Sign Flip}
\label{sec:noise}

\begin{figure}[t]
  \centering
  \includegraphics[width=\columnwidth]{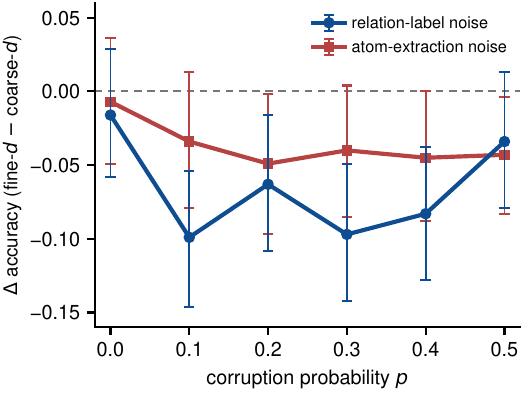}
  \caption{\textbf{No tested noise setting gives fine-$d$ a win.} Fine-$d$ $-$ coarse-$d$ accuracy on strict reverted-revert under two corruption axes (relation-label noise; atom-extraction dropout), five answer families, $n{=}445$/level, with $95\%$ CI error bars. Negative values mean coarse-$d$ wins; the point estimates remain below zero at every tested level and no setting yields a significant fine-$d$ win (some CIs cross $0$, i.e.\ ties). The relation-label curve is also non-monotone, falsifying the conjectured monotone sign flip (C3) on the axis it names.}
  \label{fig:noise}
\end{figure}

The empirical pivot of the granularity question is the fine-$d$ versus coarse-$d$ gap. We write this gap as $\Delta_{\rl-\text{Graphiti}}$. A tempting reading is that this gap sign-flips as the signal gets noisier --- fine-$d$ behind on the clean primary slice ($-0.084$), ahead on the noisier auxiliary split ($+0.090$). The checks below reject that as a clean noise effect.

The $p{=}0$ rows below are baselines for the perturbation sweep, not a second estimate of the headline clean-slice gap. The headline $-0.084$ aggregates the original primary reverted-revert comparison ($n_{\text{paired}}{=}439$), whereas each sweep rebuilds both arms on the paired noise-coverage pool with complete atom streams and five-family coverage ($89$ questions $\times$ five families, $n{=}445$). The relevant test is therefore within-sweep: as $p$ increases, does the gap move monotonically toward and across zero?

\textit{Observational precursor (confounded).} Binning all primary items by stream length --- an evidence-revision complexity proxy --- does recover a monotone trend (Table~\ref{tab:stream-length-signflip}): $\Delta_{\rl-\text{Graphiti}}$ rises from $-0.204$ on the shortest quintile to $+0.023$ on the longest. But stream length is not a pure noise variable; it co-varies with relation-type mix, topic, and item difficulty, so this is suggestive at best.

\textit{Controlled test (the noise is the only varying axis).} We therefore inject relation-scoring noise directly: holding dataset (primary reverted-revert), judge (DeepSeek), and answer model fixed, we wrap the relation function so that each predicted label is corrupted to a uniformly random other label with probability $p$, applied \emph{identically} to the fine-$d$ ledger and the coarse-$d$ arm (seeded per atom-pair so both arms see the same corruption). Table~\ref{tab:noise-curve} sweeps $p\in\{0,0.1,\dots,0.5\}$ pooled over five answer-LLM families. At clean signal within this paired sweep ($p{=}0$), the two architectures tie ($\Delta=-0.016$, n.s.) --- consistent with Proposition~2b's $b_{\text{type}}=0$ --- and every positive noise level leaves coarse-$d$ ahead. The curve is neither monotone in $p$ nor sign-flipping, so C3 fails on both parts of its monotone-crossover claim. We conclude the auxiliary crossover was driven by the confounds that distinguish the two constructions (annotator identity, relation-type mix, construction procedure), not by scoring noise, and we do not claim a controlled regime in which fine-$d$ overtakes coarse-$d$.

\textit{Extraction-noise axis.} A null on the relation-scoring axis does not by itself exclude an \emph{extraction}-noise regime favouring fine-$d$. We test that axis too: dropping each atom independently with probability $p$ (recall loss, the dominant extractor failure --- our measured event recall is $\sim$0.86), identically in both arms, then rebuilding fine-$d$ and coarse-$d$ on the surviving atoms. The atom-dropout sweep in Table~\ref{tab:noise-curve} gives the same sign conclusion: coarse-$d$ leads or ties at every tested level, with no level where fine-$d$ overtakes. Neither of the two natural noise processes tested here produces the proposed crossover.

\begin{table}[h]
\centering
\scriptsize
\setlength{\tabcolsep}{4pt}
\begin{tabular}{@{}rlrr@{}}
\toprule
Bin & Stream-length range & $n_{\text{paired}}$ & $\Delta_{\rl-\text{Graphiti}}$ \\
\midrule
1 & 2--5   & 2{,}858 & $-0.204$ \\
2 & 5--6   & 2{,}896 & $-0.168$ \\
3 & 6--10  & 2{,}875 & $-0.150$ \\
4 & 10--20 & 2{,}591 & $-0.055$ \\
5 & 20--70 & 2{,}392 & $+0.023$ \\
\bottomrule
\end{tabular}
\caption{\emph{Observational precursor, confounded.} Stream-length bins recover a monotone $\Delta_{\rl-\text{Graphiti}}$ trend, but stream length co-varies with difficulty and relation mix (text).}
\label{tab:stream-length-signflip}
\end{table}

\begin{table}[h]
\centering
\scriptsize
\setlength{\tabcolsep}{4pt}
\resizebox{\columnwidth}{!}{%
\begin{tabular}{@{}rccccc@{}}
\toprule
& \multicolumn{2}{c}{Relation-label noise} & \multicolumn{2}{c}{Atom-dropout noise} & \\
\cmidrule(lr){2-3}\cmidrule(lr){4-5}
Noise $p$ & $\Delta$ & $95\%$ CI & $\Delta$ & $95\%$ CI & $n$ \\
\midrule
0.0 & $-0.016$ & $[-0.058,+0.029]$ & $-0.007$ & $[-0.049,+0.036]$ & 445 \\
0.1 & $-0.099$ & $[-0.146,-0.054]$ & $-0.034$ & $[-0.079,+0.013]$ & 445 \\
0.2 & $-0.063$ & $[-0.108,-0.016]$ & $-0.049$ & $[-0.097,-0.002]$ & 445 \\
0.3 & $-0.097$ & $[-0.142,-0.049]$ & $-0.040$ & $[-0.085,+0.004]$ & 445 \\
0.4 & $-0.083$ & $[-0.128,-0.038]$ & $-0.045$ & $[-0.088,+0.000]$ & 445 \\
0.5 & $-0.034$ & $[-0.079,+0.013]$ & $-0.043$ & $[-0.083,-0.004]$ & 445 \\
\bottomrule
\end{tabular}
}
\caption{\emph{Controlled noise sweeps} (DeepSeek judge), pooled over five answer-LLM families ($n{=}445$ paired observations per level). Noise at level $p$ is injected identically into the fine-$d$ and coarse-$d$ arms. Relation-label noise is non-monotone and never crosses zero; atom dropout likewise gives no tested level where fine-$d$ wins.}
\label{tab:noise-curve}
\end{table}

\section{External Replication on TempLAMA (Wikidata Gold, No LLM Judge)}
\label{sec:templama}
To reduce the core findings' dependence on the self-built, LLM-labelled ESR-Bench, we replicate on TempLAMA \citep{dhingra2022templama}, a public benchmark of Wikidata $(\text{subject},\text{relation},\text{object},\text{year})$ tuples. Each $(\text{subject},\text{relation})$ is a genuine entity-state revision stream with \emph{canonical Wikidata gold}; we build streams and queries directly from the tuples (no atom extractor) and score by normalized exact match on the gold name, with \textbf{no self-built labels and no LLM judge anywhere in this experiment}. From $5{,}839$ streams we use the $237$ reverted-revert streams (a value returns) for the snapshot test and $900$ multi-value streams for provenance, over the same five answer-LLM families.

\textit{Provenance retention replicates; the residual is presentation, not typing.} Asking the value in force immediately before the current one (gold $=$ the prior distinct value), a discarding coarse-$d$ falls to floor accuracy at $0.061$, while retaining the prior state recovers much of the missing answerability: $d$-blind $0.329$, retaining coarse-$d$ $0.576$, fine-$d$ $0.902$. The retaining coarse-$d$ store, not the fine ledger, is the source of the retention threshold: it closes $+0.514$ $[+0.499,+0.531]$ over the discarding store ($n{=}4{,}500=900$ streams $\times$ five families). The remaining fine-over-retain gap is $+0.326$ $[+0.310,+0.342]$, but it cannot be credited to relation labels: both retaining renders expose the same retained states, and their residual difference is ordering/layout. The residual is larger than on ESR-Bench because TempLAMA streams are longer (often $\ge 3$ distinct values), so sorting unordered closed edges is harder than reading a pre-ordered history. The retention \emph{threshold} (discarding the prior is fatal; retention is necessary) and the \emph{representation-dominates-mechanism} reading therefore hold on external KG-gold data with exact-match scoring. The magnitude of the ordering residual is stream-length-dependent, so ESR-Bench's ``coarse retention captures \emph{most} of it'' ($+0.111$) is the short-stream case, not a universal scope point.

\textit{The snapshot render confound is not testable here (ceiling, not refutation).} On the reverted-revert streams, TempLAMA is a ceiling case rather than a mechanism test: all three renders score $\ge 0.96$ (flat chronological list $0.965$, render-only ledger $0.957$, deprecation-marked ledger $1.000$; render term $-0.008$ $[-0.021,+0.006]$, $n{=}1{,}185=237$ streams $\times$ five families), so the apparent $+0.043$ fine-over-render-only difference is not a measurable mechanism effect. The reason is simple: a clean $(\text{year},\text{value})$ tuple list makes the current value trivially readable in \emph{any} layout. This is consistent with the render confound being a property of \emph{prose-like, hard-to-read} streams (issue threads, revision histories), where layout determines whether the answer LLM can locate the current state at all; on a benchmark with no reading difficulty there is no render gap to measure. We therefore do not claim the render confound replicates on TempLAMA, only that the retention and representation-dominates findings do.

\section{Auxiliary Directional Checks}
\label{sec:audit}

The table below summarizes auxiliary directional checks on the high-noise split and external probes. These checks provide descriptive support rather than primary evidence, and should not be read as prospective evidence for Proposition~4. Their ``\rl{} $>$ \grw'' entries are \emph{not} render-matched: per the decomposition of \S\ref{sec:revrev}, such gaps over a $d$-blind baseline are largely render readability rather than the fine-$d$ mechanism, so these rows should be read as evidence that \emph{some} $d$-awareness helps (P1), not that fine typing is responsible.

We treat agreement of \emph{direction} as confirmation; magnitudes are reported but not always predicted. Rows R2, R3, and R6 all derive from Proposition~1 on overlapping data, so the seven rows test roughly five independent constructs: the $L_1$ near-null (R1), Proposition~1's separation (R2/R3/R6), the latest-known tie (R4), entity-keyed-memory failing on $L_2$ (R5), and the out-of-tier floor (R7).

\begin{table*}[t]
\centering
\scriptsize
\setlength{\tabcolsep}{4pt}
\begin{tabular}{@{}lp{1.45in}p{0.85in}p{1.4in}p{1.55in}c@{}}
\toprule
Row & Prediction & Direction & Test & Observed & Status \\
\midrule
R1 & no useful ledger gain on single-passage conflicts & near-null & FaithEval-inc., $n{=}7{,}500$ & small loss: $\Delta\!=\!-0.026\,[-0.039,-0.013]$ & $\checkmark_{\textsf{small}}$ \\
R2 & time-only entity memory fails on reverted-revert & \rl{} $>$ \grw{} & Aux. GitHub, $n{=}734$ & $+0.131\,[+0.101,+0.161]$ & $\checkmark$ \\
R3 & mixed-gold calibration trap is amplified in the auxiliary construction & \rl{} $>$ \grw{} & ESR-GitHub 5-family, $n{=}5{,}990$ & $+0.375\,[+0.362,+0.389]$ & $\checkmark_{\textsf{dir}}$ \\
R4 & latest-known streams need no ledger & \rl{} $\lesssim$ latest-only & DyKnow, $n{=}120$/family & $\varepsilon\!\approx\!0.3$ noise floor (4 fam.) & $\checkmark_{\textsf{qual}}$ \\
R5 & entity-keyed memory fails cross-source conflict & graphrag-lite $\ll$ all & WikiContradict, $n{=}1{,}265$ & 0\% recall every family & $\checkmark$ \\
R6 & held-out L3 reverted-revert separation replicates & \rl{} $>$ \grw{} & GitHub$\times$5 $+$ Wiki$\times$5, $n{=}4{,}110$ & $+0.087\,[+0.075,+0.098]$ & $\checkmark$ \\
R7 & fine ledger has out-of-tier implementation cost & $\varepsilon\!>\!0$ qualitative & DyKnow/Wiki-mono/GH-mono/LME & $\varepsilon \approx 0.3$--$0.65$ (all $\gg 0$) & $\checkmark_{\textsf{qual}}$ \\
\bottomrule
\end{tabular}
\caption{Auxiliary directional checks on the high-noise split and external probes. Confirmation is by \emph{direction} and evidence strength, not magnitude. Status legend: $\checkmark$ is a full directional match; $\checkmark_{\textsf{small}}$ is a practical near-null rather than statistical equality; $\checkmark_{\textsf{dir}}$ is direction-only, with magnitude shaped by the auxiliary construction; and $\checkmark_{\textsf{qual}}$ is qualitative. This table is separate from the primary results and from Proposition~4's channel explanation of the new reversals.}
\label{tab:audit}
\end{table*}

\section{Reproducibility}
\label{sec:repro}

\paragraph{Render-matched protocol.} The render-matched control is specified as a benchmark-agnostic transformation that other memory comparisons can reuse: given a system's structured render, construct a matched render-only ablation by forcing every co-keyed relation to \textsc{same-state} while holding layout and IDs fixed, then compare paired outputs under the same judge. This isolates layout from mechanism. The render-matched decomposition (\S\ref{sec:revrev}), controlled noise sweeps (\S\ref{sec:noise}), and provenance experiments (\S\ref{sec:provenance}) use the five answer-LLM families listed below and the same paired-scoring conventions as the corresponding main analyses.

\paragraph{Robustness controls.} All semantic-correctness numbers use a \textbf{DeepSeek} non-thinking judge as the primary axis (DeepSeek-V3 for the main analyses and render decomposition; DeepSeek-V4-flash for the provenance, noise, strata, and extraction experiments). We include three robustness controls. (i) \emph{Judge independence:} the render decomposition and the provenance gap are reproduced by an \emph{out-of-family} MiniMax-M2.5 judge; the $d$-mechanism residual stays at $+0.021$ (n.s.) and the provenance ordering is unchanged. (ii) \emph{External benchmark:} the retention ordering replicates on TempLAMA with Wikidata gold and exact-match scoring (\S\ref{sec:templama}), while the small DyKnow provenance slice concurs. (iii) \emph{Fixed analysis rules:} the external replication and deployment-fallback control used predictions and interpretation rules fixed before judging; the deployment-fallback outcome is interpreted under that fixed rule.

\paragraph{Reproducibility details.} Generation models: Llama-3.1-8B-Instruct, Mistral-7B-Instruct-v0.3, Qwen2.5-7B-Instruct, Qwen3-8B, Qwen3-14B; decoding $T{=}0$, single generation per item, batched on vLLM. DeepSeek-V3 / DeepSeek-V4-flash judge evaluations use $T{=}0$. The BERT pair scorer uses \texttt{bert-base-uncased} ($\sim$110M parameters), trained on the $500$-pair DeepSeek-labelled audit set with $5$-fold stratified CV, AdamW learning rate $2{\times}10^{-5}$, and $8$ epochs by default. The released artifacts document bootstrap seeds and $B$ values.

Data licences follow the upstream sources: GitHub issue threads under each repo's licence (the dominant \texttt{huggingface/datasets} subset is Apache-2.0), Wikipedia revisions under CC-BY-SA-4.0, and the external probes (DyKnow, LongMemEval, WikiContradict, FaithEval, StreamingQA) under their respective research-use distributions. The released artifacts include derived experiment outputs, generation and judge prompts, model versions and dates, decoding parameters, bootstrap seeds, pair-scorer metadata, and agreement-filtering rules, with upstream raw thread/revision text excluded. No raw thread/revision text or personally-identifying data is redistributed; only event IDs, atom-level extractions, and source URLs are included after a PII/credential scan.

\paragraph{Extractor coverage.} We have no human atom labels, but the benchmark's gold supporting/deprecated event ids give two automatic coverage checks. \emph{Event recall} --- the fraction of gold-relevant events that produced $\ge 1$ atom --- is $0.86$ on ESR-Bench-GitHub ($1770/2055$) and $0.84$ on MultiRepo ($108/128$); roughly one in seven gold-relevant events is missed, so the reported gains are conservative with respect to extraction recall. \emph{Co-key recall} --- among items with both a gold supporting and a gold deprecated event extracted, the fraction whose atoms land on a \emph{shared} $(x,a)$ key (the precondition for a supersession edge to fire) --- is $0.96$ on GitHub ($585/609$) and $1.00$ on MultiRepo ($38/38$). The bottleneck is therefore event-level extraction recall, not co-keying: once both endpoints are extracted they almost always reach the same ledger key, so supersession is rarely lost to mis-keying. (Mean $2.6$ atoms/event, $3.7$--$4.2$ distinct keys/stream --- the extractor is not collapsing the stream onto one key.)

\section{Supporting Noisy-Regime Tables}
\label{sec:aux-results}

Unless explicitly marked as primary, the following appendix tables are measured on the auxiliary high-noise ESR-Bench construction or on external probes. They provide the supporting details behind the directional checks in \S\ref{sec:audit}; the main text uses the DeepSeek-labelled primary surface. These auxiliary numbers are deliberately not render-matched, and several sit in a high-noise regime where \rl{} can lose on monotonic or csc through implementation cost rather than through a failure of the primary reverted-revert claim. FaithEval is the $L_1$ no-effect probe in Table~\ref{tab:audit}, so near-zero or negative \rl{} deltas there are expected. The reverted-revert slice in these tables is the larger auxiliary split ($n=734$/family), distinct from the primary DeepSeek surface.

\begin{table}[h]
\centering
\scriptsize
\setlength{\tabcolsep}{3pt}
\resizebox{\columnwidth}{!}{%
\begin{tabular}{lrrrr}
\toprule
System & Mono ($n=387$) & Refines ($n=35$) & Reverted ($n=734$) & Conflict ($n=42$) \\
\midrule
\rl{} & 0.543 & 0.086 & 0.238 & 0.643 \\
\sys{GraphRAG-lite} & 0.693 & 0.343 & 0.180 & 0.000 \\
\grw{} & 0.548 & 0.314 & 0.136 & 0.857 \\
\sys{RAG-top-k} & 0.101 & 0.029 & 0.052 & 0.500 \\
\sys{Direct context} & 0.101 & 0.029 & 0.078 & 0.452 \\
\bottomrule
\end{tabular}
}
\caption{Per-phenomenon correctness under the auxiliary Qwen-judge evaluation on the high-noise ESR-Bench-GitHub split. \rl{} wins the target reverted-revert slice while entity-keyed memory wins monotonic and refines: the load-bearing claim is structural, not uniform. Conflict is the csc boundary case: under this evaluation's gold-abstain scoring, prompt-only \grw{} abstains more readily and scores higher, GraphRAG-lite floors because it never abstains, and the rule-scored \rl{} unresolved path is the brittle decoder case discussed in \S\ref{sec:audit}.}
\label{tab:per-phenom}
\end{table}

\begin{table}[h]
\centering
\scriptsize
\setlength{\tabcolsep}{3pt}
\resizebox{\columnwidth}{!}{%
\begin{tabular}{lrrrrr}
\toprule
Family & RL cal. & GRW cal. & $\Delta$ & RL $\kappa$ & GRW $\kappa$ \\
\midrule
Llama-3.1-8B & 0.910 & 0.466 & $+0.444$ & 0.297 & 0.038 \\
Mistral-7B & 0.871 & 0.276 & $+0.594$ & 0.269 & 0.023 \\
Qwen2.5-7B & 0.634 & 0.362 & $+0.272$ & 0.089 & 0.028 \\
Qwen3-8B & 0.960 & 0.918 & $+0.042$ & 0.097 & 0.101 \\
Qwen3-14B & 0.932 & 0.407 & $+0.525$ & 0.113 & 0.042 \\
\bottomrule
\end{tabular}
}
\caption{Per-family abstain calibration and chance-corrected agreement on ESR-Bench-GitHub. Qwen3-8B has the smallest calibration gap and is the only family where \grw{} $\kappa$ matches or slightly exceeds \rl{} $\kappa$, consistent with the noise-floor saturation regime where structured-render extraction is the binding constraint (\S\ref{sec:framework}).}
\label{tab:per-family}
\end{table}

\begin{table}[h]
\centering
\scriptsize
\setlength{\tabcolsep}{3pt}
\begin{tabular}{@{}llcccc@{}}
\toprule
Family & Slice & \textsc{comp.} & \textsc{verb.} & \textsc{min.} & \textsc{json} \\
\midrule
Llama-3.1-8B  & RevRev    & $+0.046$ & $+0.119$ & $+0.155$ & $+0.075$ \\
              & FaithEval & $+0.017$ & $+0.027$ & $+0.011$ & $-0.013$ \\
\midrule
Mistral-7B    & RevRev    & $+0.170$ & $+0.253$ & $+0.173$ & $+0.215$ \\
              & FaithEval & $-0.021$ & $-0.018$ & $-0.012$ & $+0.000$ \\
\midrule
Qwen2.5-7B    & RevRev    & $+0.041$ & $+0.089$ & $+0.099$ & $+0.072$ \\
              & FaithEval & $-0.161$ & $-0.157$ & $-0.146$ & $-0.142$ \\
\midrule
Qwen3-8B      & RevRev    & $+0.158$ & $+0.289$ & $+0.257$ & $+0.312$ \\
              & FaithEval & $-0.043$ & $-0.043$ & $-0.033$ & $-0.039$ \\
\midrule
Qwen3-14B     & RevRev    & $+0.053$ & $+0.112$ & $+0.121$ & $+0.108$ \\
              & FaithEval & $-0.018$ & $-0.016$ & $-0.024$ & $-0.003$ \\
\bottomrule
\end{tabular}
\caption{Render-format ablation, full per-family $\Delta$ grid. RevRev rows are $\Delta_{\textsc{rl}-\textsc{grw}}$ on the reverted-revert slice ($n=734$/family); FaithEval rows are $\Delta_{\textsc{rl}-\textsc{rag}}$ on FaithEval-inconsistent ($n=1{,}500$/family). FaithEval is an $L_1$ intra-passage probe, so near-zero-to-negative rows are the expected no-effect pattern rather than a fine-ledger failure. All six arms use the same evaluation setup and DeepSeek-V3 judge. Qwen3 families used \texttt{/no\_think}; the Qwen3-8B RevRev sign should therefore be read under the reader-capability scope condition, not as a contradiction of the main results.}
\label{tab:render-full}
\end{table}

\begin{table}[h]
\centering
\scriptsize
\setlength{\tabcolsep}{4pt}
\begin{tabular}{@{}lrrr@{}}
\toprule
Family & \multicolumn{2}{c}{MultiRepo $\Delta_{\textsc{rl}-\textsc{grw}}$} & DyKnow \\
\cmidrule(lr){2-3}\cmidrule(lr){4-4}
       & RevRev ($n{=}43$) & full ($n{=}77$) & $\varepsilon$ ($n{=}120$) \\
\midrule
Llama-3.1-8B & $+0.116$ & $-0.065$ & --- \\
Mistral-7B   & $+0.140$ & $-0.013$ & $0.317$ \\
Qwen2.5-7B   & $+0.093$ & $-0.052$ & $0.250$ \\
Qwen3-8B     & $+0.233$ & $+0.013$ & $0.350$ \\
Qwen3-14B    & $+0.000$ & $-0.091$ & $0.241$ \\
\midrule
mean         & $+0.116$ (4/5${>}0$) & $-0.04$ & $0.290$ \\
\bottomrule
\end{tabular}
\caption{Auxiliary held-out MultiRepo and DyKnow evaluations under the same auxiliary setup, extending single-family checks to five answer families. \textbf{MultiRepo, RevRev subset ($n=43$/family):} $\Delta_{\textsc{rl}-\textsc{grw}}$; \rl{} $>$ \grw{} in 4 of 5 families (Qwen3-14B at the noise floor), cross-family mean $+0.116$, supporting P1 on a held-out repo distribution. \textbf{MultiRepo, full set ($n=77$/family):} mean $-0.04$, near zero with a slight negative tilt because the non-revrev items are mostly monotonic/refines, where the auxiliary out-of-tier analysis expects little benefit. \textbf{DyKnow:} per-family $\varepsilon$ from the auxiliary multi-family held-out evaluation; this auxiliary setting has no matched Llama-family DyKnow cell, unlike the primary DyKnow coverage in Table~\ref{tab:eval}. For the four covered families, $\varepsilon \in [0.24, 0.35]$, mean $0.29$, with magnitude left to the implementation-cost terms.}
\label{tab:aux-breadth}
\end{table}

\begin{table}[h]
\centering
\scriptsize
\setlength{\tabcolsep}{3pt}
\begin{tabular}{@{}llccc@{}}
\toprule
Family & Wording & \textsc{grw} cal. & abstain rate & $\Delta_{\textsc{rl}-\textsc{grw}}$ \\
\midrule
\multirow{3}{*}{Llama-3.1-8B} & terse         & $0.285$ & $0.750$ & $+0.625$ \\
                              & verbose (main) & $0.451$ & $0.574$ & $+0.459$ \\
                              & conservative  & $0.729$ & $0.286$ & $+0.181$ \\
\midrule
\multirow{3}{*}{Mistral-7B}   & terse         & $0.572$ & $0.452$ & $+0.299$ \\
                              & verbose (main) & $0.409$ & $0.624$ & $+0.462$ \\
                              & conservative  & $0.421$ & $0.613$ & $+0.450$ \\
\bottomrule
\end{tabular}
\caption{Robustness of the \grw{} control's calibration to the abstain-prompt wording (gold abstain rate $0.035$). Three paraphrases of the abstain instruction (terse, the paper's wording, and a conservative ``prefer to answer'' variant) move the control's calibration over a substantial band. \rl{}'s calibration is reused from Table~\ref{tab:per-family} because the ledger decision is deterministic under this wording change. \rl{} beats \grw{} in all six (variant, family) cells; the structural ordering ``selective abstain $>$ prompt-only abstain'' is invariant to the control wording, while the magnitude moves.}
\label{tab:abstain-robust}
\end{table}

\begin{table}[h]
\centering
\scriptsize
\setlength{\tabcolsep}{3pt}
\begin{tabular}{@{}llrrrc@{}}
\toprule
Family & Phenomenon & $n$ & $\Delta$ & $p$ & Holm \\
\midrule
\multirow{4}{*}{Llama-3.1-8B} & reverted-revert & 734 & $+0.131$ & $0.0001$ & \checkmark \\
 & csc (gold = \abstain) & 42 & $-0.190$ & $0.021$ & $\times$ \\
 & refines & 35 & $-0.200$ & $0.067$ & $\times$ \\
 & monotonic & 387 & $+0.003$ & $1.0$ & $\times$ \\
\midrule
\multirow{4}{*}{Mistral-7B} & reverted-revert & 734 & $+0.169$ & $0.0001$ & \checkmark \\
 & monotonic & 387 & $-0.233$ & $0.0001$ & \checkmark$_{\textsf{loss}}$ \\
 & csc (gold = \abstain) & 42 & $-0.167$ & $0.016$ & \checkmark$_{\textsf{loss}}$ \\
 & refines & 35 & $-0.057$ & $0.75$ & $\times$ \\
\bottomrule
\end{tabular}
\caption{Auxiliary high-noise RL vs \grw{} per-phenomenon results on Llama and Mistral $\times$ GitHub under DeepSeek-V3. Two-sided permutation $p$ over paired binary outcomes ($B=10{,}000$); Holm-Bonferroni at $\alpha{=}0.05$ over four tests per family. \checkmark${}_{\textsf{loss}}$ marks a Holm-significant effect in the expected-loss direction. The magnitude on monotonic and csc varies between Llama (effective tie) and Mistral (significant loss), showing the unpinned $\varepsilon$ implementation-cost term in the auxiliary noisy regime.}
\label{tab:holm}
\end{table}

\section{Operationalizing the Diagnostic (Post Hoc)}
\label{sec:tier-classifier}

The diagnostic claim --- that the conflict tier is largely a structural property of the data --- is testable post hoc: train a classifier on ESR-Bench-GitHub phenomenon labels and evaluate it held-out on ESR-MultiRepo (a different repo distribution: PyTorch, Rust, Transformers, TensorFlow). We compare a \emph{structural diagnostic} classifier against a lexical TF--IDF$+$LinearSVC over the stream text and a majority-class baseline.

The structural classifier uses nine features from extracted atoms, event metadata, and benchmark labels: event and relevant-atom counts, co-keyed value diversity, value/polarity reversal, granularity refinement, gold abstention, source-role diversity, any value change, and deprecated-event count. Because this feature set includes benchmark-side fields such as deprecated-event count and gold abstention, the table is a sanity check on structural separability, not a deployable pre-architecture router.

\begin{table}[h]
\centering
\scriptsize
\setlength{\tabcolsep}{2.5pt}
\begin{tabular}{@{}lcc|cccc@{}}
\toprule
Classifier & acc & m-F1 & RR & Mono & CSC & Ref \\
\midrule
Majority-class & $0.558$ & $0.179$ & --- & --- & --- & --- \\
Lexical (TF--IDF) & $0.766$ & $0.399$ & $0.887$ & $0.711$ & $0.000$ & $0.000$ \\
\textbf{Structural} & $\mathbf{0.818}$ & $\mathbf{0.794}$ & $0.816$ & $0.958$ & $1.000$ & $0.400$ \\
\bottomrule
\end{tabular}
\caption{Conflict-tier classification, trained on ESR-Bench-GitHub phenomenon labels and evaluated held-out on ESR-MultiRepo ($n=77$; tiers RR\,$=$\,reverted-revert, Mono\,$=$\,monotonic, CSC\,$=$\,cross-source-conflict, Ref\,$=$\,refines). The structural diagnostic recovers the tier at $0.79$ macro-F1, far above the majority baseline ($0.18$) and lexical baseline ($0.40$). RR, Mono, and CSC are recovered cleanly ($\mathrm{F1}=0.82/0.96/1.00$); the low-sample boundary is \textsc{refines} ($\mathrm{F1}=0.40$, $n=5$ test, $n=35$ train). The main confusion is RR versus Ref: 11 of 43 RR test items are predicted as Ref, reflecting overlap between value-reversal and granularity-refinement cues. Because this diagnostic uses benchmark-side fields, it is not claimed as a deployable router.}
\label{tab:tier-clf}
\end{table}

\end{document}